\documentclass{article} \PassOptionsToPackage{numbers, sort}{natbib}

\newif\ifarXiv
\arXivfalse

\usepackage{iclr2026_conference,times}
\usepackage{fontenc}
\usepackage{titletoc}

\author{
Juan Amboage$^1$, Ernst R\"oell$^{1,2,3}$, Patrick Schnider$^{4,5}$, Bastian Rieck$^{1,2}$\\[+0.5cm]
$^1$AIDOS Lab, University of Fribourg, Switzerland\\
$^2$Institute of AI for Health, Helmholtz Munich, Germany\\
$^3$Technical University of Munich, Germany\\
$^4$Department of Computer Science, ETH Zurich, Switzerland\\
$^5$Department of Computer Science, University of Basel, Switzerland
}

\usepackage{graphicx}
\usepackage{amsfonts}       \usepackage{amsmath}
\usepackage{amssymb}
\usepackage{amsthm}
\usepackage{bm}
\usepackage[english, provide=*]{babel}
\usepackage{hyperref}
\usepackage{url}
\usepackage{booktabs}
\usepackage{booktabs}       \usepackage{multicol}       \usepackage{multirow}       \usepackage{microtype}
\usepackage{comment}
\usepackage[table]{xcolor}
\usepackage[capitalise, nameinlink, noabbrev]{cleveref}
\usepackage{subcaption}
\usepackage{tabularx}
\usepackage{siunitx}
\usepackage{paralist}       \usepackage[inline]                          {enumitem}
\usepackage[hang, flushmargin]{footmisc} \usepackage{thm-restate}
\usepackage{xspace}
\usepackage{wrapfig}

\usepackage[dvipsnames]{xcolor}

\hypersetup{
    colorlinks = true,
    urlcolor = RoyalBlue,
    citecolor = RoyalBlue,
    linkcolor = RoyalBlue,
}

\declaretheorem[name=Remark]{remark}
\declaretheorem[name=Definition]{definition}
\declaretheorem[name=Theorem]{theorem}

\def\eqref#1{equation~\ref{#1}}

\def\1{\bm{1}}

\def\vh{{\bm{h}}}

\def\vp{{\bm{p}}}
\def\vq{{\bm{q}}}

\def\vv{{\bm{v}}}

\def\mA{{\bm{A}}}

\def\mD{{\bm{D}}}

\def\mI{{\bm{I}}}

\def\mL{{\bm{L}}}
\def\mM{{\bm{M}}}

\def\mQ{{\bm{Q}}}

\def\mW{{\bm{W}}}

\DeclareMathAlphabet{\mathsfit}{\encodingdefault}{\sfdefault}{m}{sl}
\SetMathAlphabet{\mathsfit}{bold}{\encodingdefault}{\sfdefault}{bx}{n}

\def\gG{{\mathcal{G}}}

\definecolor{bestgreen}{RGB}{30,200,80}

\definecolor{secondgreen}{RGB}{100,230,140}

\colorlet{worst}{red!60!black}

\newcommand{\best}[1]{\textbf{\textcolor{bestgreen}{#1}}}
\newcommand{\secondbest}[1]{\textcolor{secondgreen}{#1}}

\newcommand{\worst}[1]{\textcolor{worst}{#1}}

\newcommand{\tb}{\color{bestgreen}}
\newcommand{\tsb}{\color{secondgreen}}
\newcommand{\tw}{\color{worst}}

\newcommand{\lect}{\ensuremath{\ell\mbox{-}\mathrm{ECT}}\xspace}
\newcommand{\leapld}{LEAP-L}
\newcommand{\leaprd}{LEAP-F}

\DeclareMathAlphabet{\mathbbold}{U}{bbold}{m}{n}
\newcommand*{\boldone}{\mathbbold{1}}

\title{LEAP: Local ECT-Based Learnable Positional Encodings for Graphs}

\ifarXiv

\makeatletter
\def\maketitle
{{\renewenvironment{tabular}[2][]{\begin{flushleft}}
			   {\end{flushleft}}
\AB@maketitle}}
\makeatother

\fi

\iclrfinalcopy

\ifarXiv
\fi
\begin{document}
\maketitle

\ifarXiv
\lhead{Under review.}
\fi

\begin{abstract}
	Graph neural networks (GNNs) largely rely on the message-passing paradigm,
	where nodes iteratively aggregate information from their neighbors. Yet,
	standard message passing neural networks (MPNNs) face well-documented
	theoretical and practical limitations. Graph positional encoding (PE) has
	emerged as a promising direction to address these limitations. The Euler
	Characteristic Transform (ECT) is an efficiently computable
	geometric–topological invariant that characterizes shapes and graphs. In this
	work, we combine the differentiable approximation of the ECT (DECT) and its
	local variant (\lect) to propose LEAP, a new end-to-end trainable local
	structural PE for graphs. We evaluate our approach on multiple real-world
	datasets as well as on a synthetic task designed to test its ability to
	extract topological features. Our results underline the potential of
	\lect-based encodings as a powerful component for graph representation
	learning pipelines.
\end{abstract}

\section{Introduction}

Graphs are the preferred modality in numerous scientific domains, permitting the study of dyadic relationships in an efficient manner.
Their broad applicability comes with several challenges that make them harder to process with standard deep learning architectures.
Among these characteristics are
\begin{inparaenum}[(i)]
	\item a mixture of geometrical information~(via node and edge features) and topological information~(via the edges),
	\item highly variable cardinalities even within the same dataset, and
	\item a lack of a canonical representation.
\end{inparaenum}
The development of suitable models is thus crucial for advancing the field of \emph{graph representation learning}.
Contemporary research largely focuses on \emph{message passing neural networks}~(MPNNs), i.e., architectures that are based on local diffusion-like concepts.
While powerful, they exhibit intrinsic limitations: For instance, MPNNs tend to lose ``signals'' in graphs of high diameter~\citep{oversmoothing1, oversmoothing2, oversquashing}, and may be incapable of efficiently leveraging substructure information~\citep{counting}.

As an alternative to MPNNs, inspired by the transformer architecture \citep{attention}, recent work focuses on \emph{positional encodings}~(PEs) and \emph{structural encodings}~(SEs) of graphs, denoting functions that assign embeddings to nodes based on locality or relational information, respectively~\citep{bench-gnns, Kreuzer21a, gps}.
Most PEs/SEs are based on \emph{either} geometrical aspects~(like coordinates, curvature, or distances) \emph{or} topological aspects~(like Laplacians or random walks), which may limit their practical expressivity. To overcome this,
we propose LEAP, a new end-to-end-trainable positional encoding that leverages \emph{both} geometry \emph{and} topology.
Being based on a local, learnable variant of the Euler Characteristic Transform~(ECT), a geometrical-topological invariant, LEAP is easy to calculate and highly expressive.

\pagebreak[4]

Our paper contains the following \textbf{contributions}:
\begin{enumerate}[leftmargin = 1em, nosep]
	\item We propose a new graph positional encoding based on local ECTs, which is highly flexible and permits end-to-end training, specifically geared to work with geometric graphs.
    
	\item We observe that our method captures structural differences in graphs even in case the node features are \emph{non-informative},
        thus also permitting to solve learning tasks for non-attributed graphs.
    
	\item We conduct extensive experiments on benchmark datasets that demonstrate that our method yields \emph{improved predictive power} in comparison to existing positional encodings when used in conjunction with graph neural networks.
\end{enumerate}

\section{Background}
Before introducing our learnable positional encoding, we provide a
short self-contained summary of message-passing, positional encoding in the context of graphs, and the Euler Characteristic Transform.

\subsection{Message Passing}

Graph Neural Networks (GNNs) are specifically designed to operate on graph-structured data. A large subclass of GNNs are Message Passing Neural Networks~\citep[MPNNs]{mpnn}.
MPNNs represent each node by a vector that is iteratively updated by aggregating  neighboring representations. Hence, the state of a node $v$ at step $t$, denoted $\vh_v^{(t)}$, is computed as
\begin{equation}
    \vh_v^{(t)} = \text{UPDATE}\left(\vh_v^{(t-1)}, \text{AGGREGATE}\left(\{\vh_u^{(t-1)} : u \in \mathcal{N}(v)\}\right)\right),
\end{equation}
where both AGGREGATE and UPDATE are learnable functions and $\mathcal{N}(v)$ denotes the neighbors of node $v$.
Following \citet{lect}, we refer to a graph $\gG$ together with feature vectors for each of its nodes as a \emph{featured graph} and adopt the following notation.

\begin{definition}
A \emph{featured graph} is a pair $(\gG, x)$, where $\gG$ is a (non-directed) graph, and  $x$ is a map that assigns each node $v \in V(\gG)$ a feature vector $x(v)\in\mathbb{R}^d$. We denote the set of nodes of $\gG$ by $V(\gG)$, and the set of edges by $E(\gG)$.
\end{definition}

Despite their popularity, common MPNNs are limited by phenomena such as oversquashing~\citep{oversquashing}, 
oversmoothing~\citep{oversmoothing1, oversmoothing2}, or restricted expressive power \citep{howpowerful, counting}. 
Multiple approaches have been proposed to address these challenges, for instance by
\begin{inparaenum}[(i)]
    \item modifying graph connectivity via virtual nodes~\citep{virtualnode1, bench-pe},
    \item combining message passing with global attention~\citep{gps}, or
    \item imbuing a model with topology-based inductive biases~\citep{tnn, Horn22a}.
\end{inparaenum}

\subsection{Graph Positional Encodings}
\label{sub:gpe}

Inspired by positional encodings in transformers~\citep{attention}, graph positional encodings emerged as a way to inject structural information into the node features. 
Architectures such as GPS~\citep{gps} combine multiple PEs, enabling global-attention layers to incorporate graph structure. Graph PEs have also been shown to benefit standard MPNNs~\citep{bench-gnns,learn-pe,gat-pe, tope}. \citet{gps} propose a categorization of graph PEs into \emph{Positional Encodings} and \emph{Structural Encodings}, further subdivided into \emph{local}, \emph{global}, or \emph{relative} variants.
Two commonly-used graph positional encodings are the Random Walk Positional Encoding~\citep[RWPE]{learn-pe} and the Laplacian Positional Encoding~\citep[LaPE]{gen-lape}, which we will briefly introduce below. 
Both methods inspired several other non-learnable approaches~\citep{gen-lape, bench-pe, signNet, gps}, as well as learnable ones~\citep[SignNet]{signNet}.

\paragraph{Random Walk Positional Encoding~(RWPE).}
For any node $v\in V(\mathcal{G})$, \citet{learn-pe} define the $k$-dimensional RWPE of $v$, denoted by $\vp_v^{\textnormal{RWPE}_k}$ as
\begin{equation}
	\vp_v^{\textnormal{RWPE}_k} := [\textbf{RW}_{vv}, (\textbf{RW})^2_{vv}, \ldots, (\textbf{RW})^k_{vv}] \in \mathbb{R}^k,
\end{equation}
where $\textbf{RW}:= \mA(\gG)\mD(\gG)^{-1}$ is the random walk matrix of the graph $\gG$, $\mA(\gG)$ denotes the \emph{adjacency matrix} of $\gG$, and $\mD(\gG)$ denotes the \emph{degree matrix} of $\gG$. \citet{gps} categorize RWPE as a \emph{local structural encoding}.

\paragraph{Laplacian Positional Encoding (LaPE).}
The \emph{normalized Laplacian matrix} of $\mathcal{G}$ is given by $\mL(\gG) = \mI - \mD(\gG)^{-1/2}\mA(\gG)\mD(\gG)^{-1/2}$,
where $\mI$ denotes the identity matrix.
The LaPEs of the nodes in $\gG$ are constructed from the eigendecomposition
of $\mL(\gG)=\mQ^\top \Lambda\ \mQ$.
Given the eigenvalues sorted in ascending order
${\lambda^{(1)},\ldots,\lambda^{(K)}}$, with corresponding
eigenvectors ${\vq^{(1)},\ldots,\vq^{(K)}}$, \citet{bench-gnns} define the
$k$-dimensional LaPE ($\vp_v^{\textnormal{LaPE}_k}$) of a node $v$ as
\begin{equation}
	\vp_v^{\textnormal{LaPE}_k} := [\vq^{(i)}_{v}, \vq^{(i+1)}_{v}, \ldots, \vq^{(i+k-1)}_{v}] \in \mathbb{R}^k,
\end{equation}
where $i$ is the index of the first non-trivial eigenvector. 
Since LaPE employs the eigendecomposition of the full graph, it is considered to be a \emph{global positional encoding}~\citep{gps} .

\subsection{The Euler Characteristic Transform~(ECT)} \label{ssec:Euler Characteristic Transform}

The \emph{Euler Characteristic Transform}~(ECT) originated as a method to summarize simplicial complexes, i.e., higher-order domains~\citep{ect}.
Being an expressive and computationally favorable summary statistic, it has found many interesting applications
in the biomedical sciences~\citep{amezquita2022measuring}, computer vision~\citep{jiang2020weighted,cisewskikehe2023weighted}, statistical functional analysis~\citep{Crawford20a}, and, more recently, in generative tasks~\citep{point-transform}. 
For an extensive exposition of the applications and a review of the literature, we refer the reader to~\citet{ect-invitation} or \citet{Rieck25}.

We will  restrict our exposition to the case of graphs, consisting of \emph{vertices} and \emph{edges}.
For graphs, the \emph{Euler characteristic} is defined as the number of nodes minus the number of edges; it is a topological 
invariant of the graph.
When the Euler characteristic of two graphs is different, the graphs are not topologically equivalent, permitting us to distinguish them.\footnote{Formally, homotopy-equivalent topological spaces have the same Euler characteristic.
}
However, as  many graphs have the same Euler characteristic, its expressive power remains limited.
By moving to a \emph{multi-scale} variant of the Euler characteristic, we obtain the ECT, which combines geometrical and topological information to obtain a more expressive representation. 
Specifically, given a featured graph $(\gG,x)$, we calculate the inner product of its attributes with a unit vector $\theta\in \mathbb{S}^{d-1}$, referred to as a \emph{direction}, and consider the pre-image of the inner product to obtain a monotonically increasing sequence of subgraphs of $\gG$.
Tracking the Euler characteristic along that sequence indexed by $t\in \mathbb{R}$ yields the \emph{Euler Characteristic Curve}~(ECC) in the direction of $\theta$.
The map that sends each direction vector to its corresponding ECC is called the \emph{Euler Characteristic Transform}~(ECT). 
For graphs, it is defined as
\begin{equation}
	\begin{aligned}
		\mathrm{ECT}\colon \mathbb{S}^{d-1}\times\mathbb{R} &\to \mathbb{Z}                                                                                \\
		(\theta,t)                                    &\mapsto  \sum_{v \in V(\gG)} \boldone_{[\langle \theta,\ x(v) \rangle,\ \infty)}(t) - \sum_{e \in E(\gG)} \boldone_{[\max_{u\in e}\langle \theta,\ x(u) \rangle,\ \infty)}(t).
	\end{aligned}
	\label{eq:ECT indicator functions}
\end{equation}
Somewhat surprisingly, given a sufficiently large \emph{finite} number of directions, the ECT is \emph{injective} on geometric graphs and geometric~(simplicial) complexes~\citep{ectdirections, inverse-ect}, i.e., distinct inputs yield distinct ECTs.
One limiting factor to the applicability of the ECT in a deep learning setting is the
lack of differentiability with respect to the direction vectors and input
coordinates.
However, by approximating the indicator function of \cref{eq:ECT indicator functions} with a sigmoid function,
we obtain the \emph{Differentiable Euler Characteristic Transform}~\citep[DECT]{dect}, which may be integrated into standard deep learning pipelines.
This formulation of the ECT provides a \emph{global} summary of a shape, but certain graph learning
tasks benefit from a \emph{local} perspective of the graph around a node of interest.
As a \emph{static}, i.e., non-trainable, extension to the ECT, the \emph{local Euler Characteristic Transform}~\citep[\lect]{lect},
constitutes a variant based on local neighborhoods with favorable properties
for node classification.
Given a featured graph $(\mathcal{G},x)$ with $x\colon V(\mathcal{G})\to\mathbb{R}^d$,
and a vertex $v$, the local ECT  of $v$ with respect
to $m\in \mathbb{N}$ is defined as
\begin{equation}
	\ell\textnormal{-ECT}_m[\gG,x;v]:=\textnormal{ECT}[\mathcal{N}_m(v,\gG), x|_{V(\mathcal{N}_m(v,\gG))}],
\end{equation}
where $\mathcal{N}_m(v,\gG)$ denotes a neighborhood of $v$,
whose locality is controlled by the hyperparameter $m$.
The following result by~\citet{lect} relates the \lect to MPNNs.

\begin{theorem}
	Let $(\gG, x)$ be a featured graph, and let $\{\ell\textnormal{-ECT}_1[\gG,x; v]\}_v$ be the set of the
	1-hop $\ell$-ECTs of all the vertices $v\in V(G)$.
Then $\{\ell\textnormal{-ECT}_1[\gG,x; v]\}_v$ provides all the (non-learnable) needed information to perform a single message passing step on $(\gG, x)$.
\end{theorem}

The required non-learnable information for a single message passing
step refers to the fact that for a given vertex $v$, one can
theoretically recover the features of the neighboring nodes from the
$\ell$-ECT.
This result highlights the power of the 1-hop $\ell$-ECT for graph
representation learning.
Moreover, \citet{lect} show that the \lect is sufficiently expressive to perform subgraph counting,
one of the limitations of traditional message passing architectures~\citep{counting}.
This illustrates that ECT-based methods can be \emph{more powerful} than traditional
message passing neural networks in certain cases. \section{Methods}
This section introduces the Local ECT and Projection PE~(LEAP), a \emph{learnable}
local structural graph PE based on the $\ell$-ECT. As part of this encoding, we
present strategies for projecting the ECT of a shape into a lower-dimensional
space.

\subsection{\texorpdfstring{$\ell$-ECT based Positional Encoding}{l-ECT based Positional Encoding}}

Given a featured graph $(\mathcal{G}, x)$ with $d$-dimensional node features,
which may be static (i.e., the original node features or another PE), or learned
(i.e., hidden states at some step of an MPNN), let $\mathbb{T} \subset [0,1]$ be
a finite set of thresholds and $\Theta \subset \mathbb{S}^{d-1}$ a finite set
of directions. The $k$-dimensional LEAP PE of a node $v \in V(\mathcal{G})$ is
constructed as follows:

\begin{enumerate}[leftmargin = 1em, nosep]
	\item Compute the $m$-hop subgraph $\mathcal{N}_m(v, \mathcal{G})$ around
	      node $v$.
	\item Given the set of nodes $\{u_1, \ldots, u_n\} = V(\mathcal{N}_m(v,
		      \mathcal{G}))$, mean-center their feature set $\{x(u_1), \ldots, x(u_n)\}$
	      and divide each element by the maximum norm in the centered set to obtain
	      new features $\mathbb{F} = \{f(u_1), \ldots, f(u_n)\} \subset
		      \mathbb{S}^{d-1}$, where $f \colon V(\mathcal{N}_m(v, \mathcal{G})) \to
		      \mathbb{F}$ denotes the mapping between each node in the $m$-hop and its
	      normalized feature vector.

	\item Compute the matrix $\mM \in \mathbb{R}^{|\Theta|\times|\mathbb{T}|}$
	      whose $(i,j)$ entry is the differentiable approximation of the ECT of
	      $(\mathcal{N}_m(v, \mathcal{G}), f)$ at $(\theta_i, t_j) \in
		      \Theta \times \mathbb{T}$.

	\item Lastly, a learnable projection $\phi:
		      \mathbb{R}^{|\Theta|\times|\mathbb{T}|} \to \mathbb{R}^k$ maps $\mM$ to a
	      vector $\text{PE}(v)\in \mathbb{R}^k$, which is the final positional
	      encoding of node $v$.
\end{enumerate}

\begin{figure}
	\centering
	\includegraphics[width=.8\linewidth]{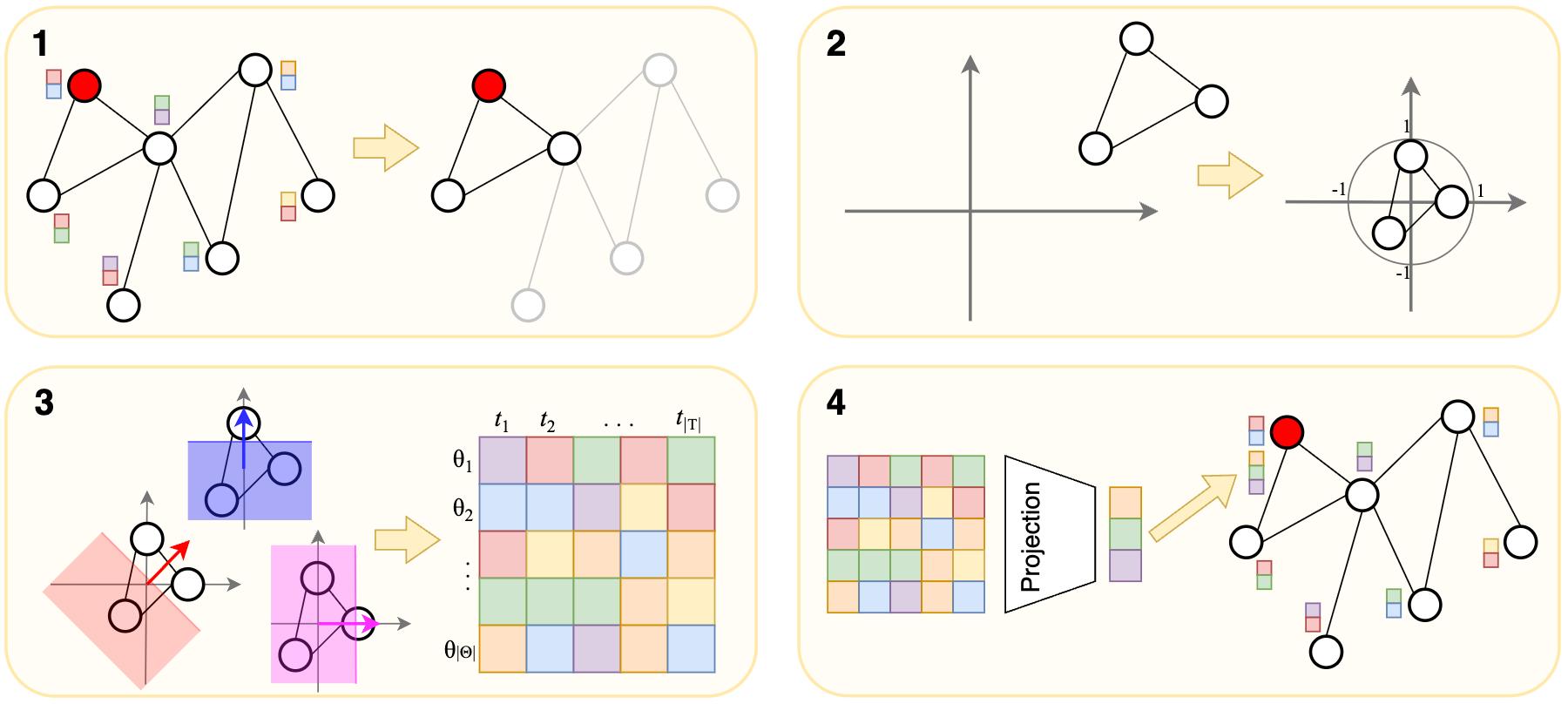}
	\caption{Steps for computing the LEAP PE using $1$-hop neighborhoods. (1)
		The neighborhood of a node in a featured graph is selected. (2) Normalization
		of the neighborhood features. (3) Computation of the differentiable ECT. (4)
		Projection of the matrix representation of the ECT to get the PE vector.
	}\label{fig:LEAP-steps}
\end{figure}

\begin{remark}
	\emph{LEAP} is \emph{not} a static pre-processing step on the graph. On the
	contrary, it can be integrated in graph neural network architectures to be trained
	in an end-to-end fashion.
\end{remark}
The previous remark highlights a key difference between LEAP and
graph PEs  like LaPE and RWPE. This is also an important
distinction from the prior use of the $\ell$\mbox{-}ECT, which was introduced as a static, non-learnable extension of node features, with neighborhood connectivity being disregarded as the ECT was calculated on node neighborhoods as if they were point clouds rather than graphs~\citep{lect}.
In addition, LEAP permits the set of directions $\Theta$ to be randomly initialized and then \emph{either} kept fixed \emph{or} optimized during training. LEAP can also be applied to learned graph features and, since it integrates with any GCN, it is naturally applicable to both graph-level and node-level tasks. By contrast, the DECT is geared towards generating graph-level descriptors~\citep{dect}.

\begin{remark}
	Within the categorization of \citet{gps}, LEAP is a \emph{local structural encoding}.
\end{remark}

The locality of LEAP comes from computing each node’s encoding only from its $m$-hop subgraph. Thus, locality is controlled by the hop number~$m$, which serves as a hyperparameter. By default, we suggest $1$-hop neighborhoods, making our method \emph{as scalable} as message passing, but we also describe two ways to control the locality of LEAP:

\begin{itemize}[leftmargin = 1em, nosep]
	\item Use a larger hop number $m$. While  straightforward, it should be noted that two nodes may differ in their $m$-hop neighborhoods while becoming identical at $(m+1)$-hops.\footnote{For sufficiently large $m$, this strategy yields identical PEs for all nodes within the same connected component.}

	\item Alternatively, we compute LEAP multiple times for each node with increasing $m$, then concatenate the results to obtain a PE that captures how the $m$-hop neighborhoods evolve as $m$ grows.
\end{itemize}

We also note that two nodes in a graph that share identical $m$-hop neighborhoods receive the same LEAP PE, since the second step in the computation of the PE yields identical outputs. This aligns directly with the definition of \emph{local structural encoding} given in \citet[Table~1]{gps}.
Moreover, consider a node whose normalized $m$-hop neighborhood features form a geometric
graph embedding.\footnote{In general, there is no guarantee this will occur.}
If we could access the ECT of that subgraph rather than an approximation, then
by the injectivity results of the ECT~\citep{inverse-ect, ectdirections} we
would have all the information required to recover the neighborhood’s
structure.\footnote{We design an experiment to test the ability of LEAP to	capture topological
	features of a graph, see \cref{subsec:syn}.
}

\subsection{ECT projection strategies}
Since LEAP aims to capture structural information, it should be invariant to
scaling and rotations of neighborhood features. Step~2 above addresses
normalization, but to minimize the effect of rotations, the projection in Step~4
should be \emph{permutation invariant} with respect to the ECCs.
However, this requirement is often ignored in practice~\citep{dect}.
In the remainder of this section, we present five
projection strategies for LEAP, some of which explicitly enforce this
invariance.

\begin{description}[leftmargin = 1em, itemsep = 0.5em, parsep = 0pt]

\item[Linear projection:]
We ``flatten'' the \lect of each node into a vector $\vv \in \mathbb{R}^{D}$ with $D =
		      |\Theta|\cdot |\mathbb{T}|$, following \citet{amezquita2022measuring}.
We then apply a linear projection by multiplying $\vv$ with a learnable
	      matrix $\mW \in \mathbb{R}^{k \times D}$.
This projection is \emph{not} permutation invariant with respect to the ECC, and the
	      number of learnable parameters with respect to $|\Theta|$ and $|\mathbb{T}|$
	      is $\mathcal{O}(|\Theta|\cdot |\mathbb{T}|)$.

\item[One-dimensional convolutions:]
We treat the \lect {} of each node as a multichannel time series, where
	      thresholds act as time steps and each ECC defines a channel. Several 1D
	      convolutions are concatenated, and the resulting channels are averaged to
	      produce a vector that is used as an input to an MLP. This projection is \emph{not} permutation
	      invariant with respect to the order of the directions,
and	the number of learnable parameters  with respect to $|\Theta|$ and $|\mathbb{T}|$ is
	      $\mathcal{O}(|\Theta|+|\mathbb{T}|)$.
\item[DeepSets:]
We treat the \lect of a node as a set of $|\mathbb{T}|$-dimensional vectors,
	      corresponding to the ECCs along different directions in $\Theta$, processing this set
	      using an architecture inspired by DeepSets~\citep{deepsets}: Given the set of
	      vectors corresponding to the ECCs we have
	      $\text{PE}= \text{MLP}_2( \sum_{\theta \in
			      |\Theta|}\text{MLP}_1(\text{ECC}_\theta))$. This projection
	      strategy is permutation invariant wrt.\ the directions of the ECT, and
	      its number of learnable parameters is independent of $|\Theta|$.

\item[Attention:]
We treat the \lect{} of a node as a set of $|\mathbb{T}|$-dimensional
	      vectors, corresponding to the ECCs along the different directions in $\Theta$, and
	      process this set by a transformer encoder with a single attention head.
To obtain the PE, we apply an MLP to the sum of the generated ECC representations.
Due to the use of a self-attention without any
	      positional encoding, the projection is permutation invariant, and the number of
	      learnable parameters depends on $|\mathbb{T}|$ but not on $|\Theta|$.

\item[Attention with PE:]
As a variant of the previous projection,
          instead of feeding the transformer encoder the set of $\text{ECCs}$ directly, we concatenate each $\text{ECC}_\theta$ with the corresponding direction $\theta \in \Theta$ before passing it to the encoder. This yields a permutation invariant projection strategy, while incorporating information about the directions along which the ECCs were computed.

\end{description}

\subsection{Properties}

    We first discuss the \emph{computational complexity} of our method. Given an $m$-hop subgraph 
    $\mathcal{N}_m(v, \mathcal{G})$ for each vertex~$v$, calculating the \lect{} has a total computational complexity of
    $\mathcal{O}(\sum_v | \mathcal{N}_m(v, \mathcal{G}) |$).
In the worst case, each subgraph is the \emph{complete} graph on $n$ vertices, leading to an 
    overall complexity of $\mathcal{O}(n^3)$. For \emph{sparse graphs}
    whose $m$-hop neighborhood is of the order of $m = \mathcal{O}(n)$, we obtain a worst-case 
    complexity of $\mathcal{O}(n^2)$.
Finally, assuming \emph{bounded degree}, this reduces to a worst-case complexity of
    $\mathcal{O}(n)$, which is asymptotically equal to one step of message passing.
Moreover, individual \lect{}s can be computed \emph{in parallel}.
In terms of expressivity, \citet{lect} provide the theoretical foundation for our work,
    stating that, given a sufficiently large number of directions, the injectivity of the \lect{} guarantees
    that it is \emph{more} expressive than message passing.
However, we consider the main contribution of our work to be the development of a novel local positional encoding
    and its empirical evaluation, in the spirit of \citet{gps}, leaving a more in-depth theoretical analysis for
    future work.

\section{Experiments}

We conduct experiments to evaluate different aspects of LEAP, investigating
\begin{inparaenum}[(i)]
	\item its ability to capture structural properties \emph{independent} of node features,
	\item its impact on the performance of different graph
neural network
    architectures and the effect of learning the directions of the transform,
\item its performance on a large-scale dataset with 202{,}579 graphs~\citep{alchemy},
	\item its behavior when applied to learned node features in the \emph{HIV} dataset~\citep{moleculenet}, and
	\item the effect of hyperparameters.
\end{inparaenum}
Subsequently, \emph{LEAP-L} indicates that the directions for LEAP were randomly
initialized and learned during training, while \emph{LEAP-F} denotes that the
directions remained fixed.

\subsection{Synthetic Dataset}
\label{subsec:syn}
We introduce a synthetic dataset of 40,000 graphs to test whether LEAP can capture structural differences \emph{independent} of node features, thus proving that LEAP is indeed a \emph{structural encoding}.
Each graph has three nodes and contains either zero, one, two, or three edges, yielding a classification task with four classes based on edge count.
The node features are uniformly sampled from the unit disk $D_1\subset \mathbb{R}^2$ to make the task
purely structural.
We use a standard GCN and GAT architectures as base models, and  compare them to the same model enhanced with LEAP added as structural
positional encoding.
For the computation of the ECT used in LEAP, we use $16$ directions with a resolution
of $16$,  summarizing each graph into a $16\times 16$ ECT.
The models enhanced with LEAP achieve a perfect accuracy of
$100.0 \pm 0.0$, demonstrating LEAP's ability to capture
structural properties \emph{independent} of the node features.
By contrast, the GCN and GAT models alone exhibited lower accuracies~($71.83 \pm 0.27$ and
$69.44 \pm 0.82$, respectively), demonstrating their inability to capture relevant
structural graph properties when informative node features are not available\footnote{Although LaPE and RWPE are computed independently of node features, we also performed this experiment using these PEs. Like LEAP, they achieved perfect accuracy.}.

\subsection{Classifying real-world datasets}
\begin{wraptable}{l}{0.5\linewidth}
	\vspace{-1.2\baselineskip}
	\centering
	\sisetup{
		detect-all              = true,
		table-format            = 2.1,
		detect-mode             = true,
		separate-uncertainty    = true,
		retain-zero-uncertainty = true,
		mode                    = text,
	}\caption{Best approach (architecture, PE strategy, and projection strategy) and
		relative accuracy improvement with respect to the worst performing baseline
		for TU classification datasets. In all cases the best result was achieved
		using our PE strategy.
	}\label{tab:best-approaches}
	\resizebox{\linewidth}{!}{\begin{tabular}{ll SS S}
			\toprule
			{\small\sc Dataset}     & {\small\sc Best Method}                & {\small\sc Worst} & {\small\sc Best} & {\small\sc Gain (\%)} \\
			\midrule
			{\sc\small Letter-H   } & {\small NoMP + \leapld + 1D Conv     } & 41.6            & 81.6             & 96.2                  \\
			{\sc\small Letter-M   } & {\small NoMP + \leapld + 1D Conv     } & 57.8            & 88.5             & 53.1                  \\
			{\sc\small Letter-L   } & {\small NoMP + \leapld + 1D Conv     } & 80.4            & 98.0             & 21.9                  \\
			{\sc\small Fingerprint} & {\small NoMP + \leapld + Linear      } & 48.8            & 55.7             & 14.1                  \\
			{\sc\small COX2       } & {\small GAT  + \leapld + Attn w/\ PE } & 77.7            & 80.1             & 3.1                   \\
			{\sc\small BZR        } & {\small NoMP + \leapld + Linear      } & 78.3            & 84.7             & 8.2                   \\
			{\sc\small DHFR       } & {\small GCN  + \leapld + Attn w/\ PE } & 70.1            & 77.6             & 10.7                  \\
			\bottomrule
		\end{tabular}
	}
\end{wraptable}
 We evaluate LEAP on several graph classification datasets from the TU benchmark \citep{tu-datasets}.
Our aim is to evaluate
\begin{inparaenum}[(i)]
	\item the capacity of LEAP to enhance existing graph neural networks with structural information,
	\item compare LEAP with existing PEs, and
	\item investigate in which architecture LEAP induces the largest increase in accuracy.
\end{inparaenum}
Of particular interest is the evaluation on the \emph{Alchemy\_full}~\citep{alchemy} dataset,
as the regression targets are rotation invariant with respect to the node features.
For this dataset, we normalize the regression targets so that all 12 tasks are on the same scale.
For the \emph{HIV} dataset \citep{moleculenet}, where nodes have categorical features, we exploit the end-to-end differentiability of the ECT by extending the architecture with a learnable embedding layer that maps these features into $\mathbb{R}^3$, where LEAP is computed.

\paragraph{Architectures.}
We fix four ``backbone'' architectures to which
we add different positional encodings, namely
\begin{inparaenum}[(i)]
	\item a GCN \citep{gcn},
	\item a GAT \citep{gat}, 
    \item a GIN \citep{howpowerful}, and
	\item NoMP~(``no message passing''), a model that we introduce based on a transformer encoder.
\end{inparaenum}
Following \citet{k-forms}, we use five message-passing layers and 32-dimensional hidden states for GCN and GAT.
For the \emph{Alchemy} dataset, we scale the architectures to $10$ layers with
$64$-dimensional hidden states.
The NoMP architecture projects node features into a $16$-dimensional latent space with a linear layer, followed by a single self-attention layer that produces a $16$-dimensional state for each node. The final classification is performed by a feedforward layer. We chose hyperparameters to match the parameter count of GCN/GAT.
By design, NoMP ignores graph structure unless given positional encodings, thus permitting us to evaluate the ability of each PE to encode relevant structural properties.
Finally, as positional encodings, we consider the following baselines:
\begin{inparaenum}[(i)]
	\item No positional encoding,
	\item RWPE, which, like LEAP, is a local structural PE under the categorization of \citet{gps}, making it a particularly relevant baseline, and
	\item LaPE, a widely used graph PE that, unlike LEAP, captures global positional information.
\end{inparaenum}

\paragraph{Experimental setup.}
All experiments use 5-fold cross-validation and are trained with the Adam optimizer
for up to $100$ epochs with early stopping enabled.
As a loss term, we use the \emph{cross entropy loss} except for the \emph{Alchemy} dataset,
where we use the mean squared error loss.
We use 10-dimensional PEs for all types and datasets. The only difference between the backbones with or without a PE is that the input dimension of the backbone increases by 10 when a PE is used.
The Euler Characteristic Transform in LEAP is calculated with $16$ directions and $16$ thresholds.
To simplify the setup, we keep all hyperparameters of LEAP's projection strategies \emph{fixed} across all datasets. Despite this, as we describe below, we observe high predictive performance across a variety of datasets.

\subsection{Results}
\begin{wraptable}[11]{l}{0.5\textwidth}
	\vspace{-1.2\baselineskip}
	\centering
	\sisetup{
		detect-all              = true,
		table-format            = 2.2(1.2),
		detect-mode             = true,
		separate-uncertainty    = true,
		retain-zero-uncertainty = true,
	}\caption{Surprisingly, increasing the neighborhood size ($\mathcal{N}_m$) does not
		improve the efficacy of LEAP, showing that the $1$-hop neighborhood
		is sufficient.
	}
	\resizebox{\linewidth}{!}{\begin{tabular}{lcSSS}
			\toprule
{\small\sc Method}          & {\sc\small $\mathcal{N}_{m}$} & {\small\sc Letter-H} & {\small\sc Letter-M} & {\small\sc Letter-L} \\
			\midrule
			\multirow[c]{3}{*}{\leaprd} & 1                             & \tb 81.29 \pm 1.91   & \tb 88.0 \pm 1.89    & \tb 96.27 \pm 0.84   \\
			                            & 2                             & 74.44 \pm 3.26       & 84.31 \pm 0.76       & 94.13 \pm 1.05       \\
			                            & 1, 2                          & 77.91 \pm 1.82       & 84.76 \pm 1.4        & 96.09 \pm 0.34       \\
			\midrule
			\multirow[c]{3}{*}{\leapld} & 1                             & \tb 80.62 \pm 3.58   & \tb 86.49 \pm 2.2    & \tb 96.18 \pm 1.24   \\
			                            & 2                             & 72.76 \pm 2.66       & 85.11 \pm 1.29       & 93.38 \pm 0.84       \\
			                            & 1, 2                          & 78.13 \pm 2.83       & 85.96 \pm 1.42       & 95.64 \pm 1.51       \\
			\bottomrule
		\end{tabular}
	}
	\label{tab:Ablation Hops}
\end{wraptable}

 \cref{tab:LEAP Best Projection} reports the results for LEAP and the baselines in
combination with different architectures across the various classification datasets.
For every dataset--architecture combination, the two LEAP variants (F/L) achieve
the best and second-best performance.
When combined with GCN and NoMP architectures, \emph{learning} the directions of LEAP
consistently improves performance in comparison to keeping them fixed.
For GAT, learning the directions slightly reduces performance compared to the fixed
variant of LEAP in 3 of the 7 datasets.
For all datasets, the overall best-performing architecture–PE combination always uses
LEAP with learned directions.

\begin{figure}[t]\begin{subfigure}[t]{0.39\linewidth}
		\centering
		\includegraphics[width=.9\linewidth]{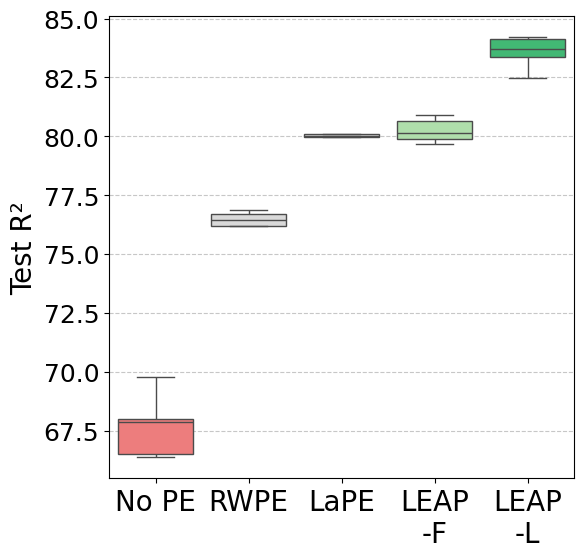}
		\caption*{\emph{Alchemy}}
	\end{subfigure}\hfill
	\begin{subfigure}[t]{0.59\linewidth}
		\centering
\includegraphics[width=.9\linewidth]{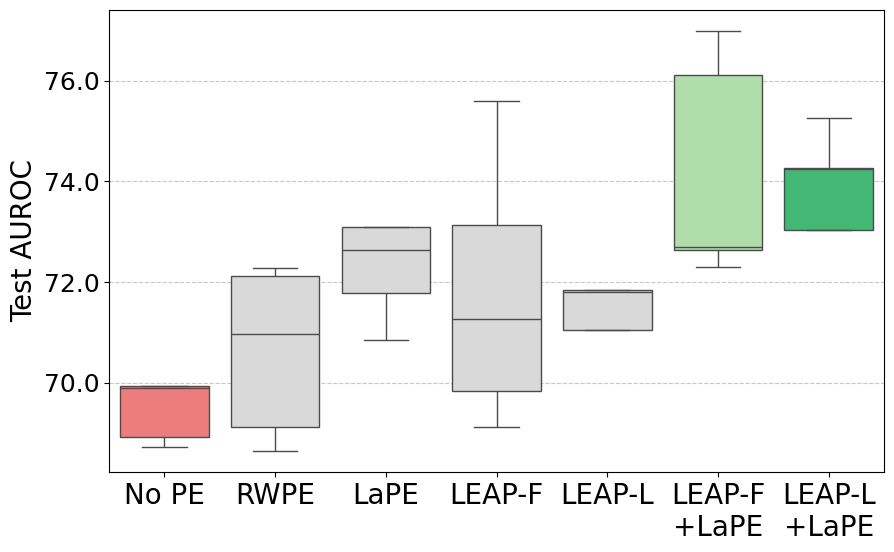}
		\caption*{\emph{HIV}}
		\label{fig:subfig1-hiv}
	\end{subfigure}\caption{Results for different PE strategies on the \emph{Alchemy} and \emph{HIV} datasets
		reporting the $R^2$ and AUROC using a GCN.
Colors rank the PEs from \best{best}, \secondbest{second best}
		to \worst{worst}.
LEAP with learnable direction significantly outperforms other methods on the
		\emph{Alchemy} dataset while performing second best on the \emph{HIV} dataset.
}\label{fig:Box Plot GCN HIV and Alchemy}
\end{figure}

\cref{tab:best-approaches} reports the relative increase of the best-performing method
compared to the worst-performing method.
We observe the largest gains from using LEAP on the \emph{Letter}
and \emph{Fingerprint} datasets.
For the datasets \emph{COX2}, \emph{BZR}, and \emph{DHFR}, the advantage over the
baselines is less pronounced, likely due to their smaller size, making it
harder to benefit from richer features. 
We observe the largest 
improvement on \emph{DHFR}, the largest dataset among the
three.
\cref{tab:LEAP Best Projection} also shows that NoMP
\emph{without} positional encodings outperforms baseline GNNs on the
\emph{Letter-High} and \emph{Letter-Low} datasets, highlighting the limitations of MPNNs, i.e., models \emph{without}
structural information may achieve better results.

We also evaluate LEAP on the \emph{Alchemy} and
\emph{HIV} datasets from the TU and MoleculeNet benchmarks, respectively, following the same setup described above.
The \emph{Alchemy} dataset is significantly larger compared to the other datasets allowing
LEAP to extract more meaningful information from the data.
For the \emph{Alchemy} dataset, we show the $R^2$ score and for the \emph{HIV} dataset, we report AUROC,
due to large class imbalances.
\cref{fig:Box Plot GCN HIV and Alchemy} shows the results for LEAP with the GCN
backbone and the various PE methods, showing a clear advantage for LEAP~(with
learned directions) over all baselines.
For the \emph{HIV} dataset, results exhibit a large degree of variability, and this is the only case where
neither of the two LEAP variant~(L/F) outperforms all baselines. Still, both
variants surpass RWPE, which is in the same category of
\emph{local structural encodings}.
LaPE yields the strongest performance for a single PE, suggesting that \emph{global	positional} information, which cannot be captured by LEAP, may be particularly relevant in this case. For this reason, we also evaluate the use of LaPE and LEAP combined through concatenation. We observe that the combination of both PEs yields the best overall results. This showcases how these two PE strategies can successfully complement each other.
Please refer to \cref{fig:Box Plot Alchemy Full} and \cref{fig:Box Plot HIV} for the full results.

\subsection{Ablations}

After having established that LEAP captures essential structural information to
be used with multiple graph neural network architectures, we further aim to
investigate the sensitivity of LEAP with respect to its various components.
In particular we hope to further understand how
\begin{inparaenum}[(i)]
	\item the choice of projection method,
	\item the number of hops, and
	\item the embedding dimension
\end{inparaenum}
impact the performance of LEAP.

\begin{wraptable}[18]{l}{.5\textwidth}
	\vspace{-1.2\baselineskip}
	\centering
	\sisetup{
		detect-all              = true,
		table-format            = 2.2(1.2),
		detect-mode             = true,
		separate-uncertainty    = true,
		retain-zero-uncertainty = true,
		mode                    = text,
	}\caption{Ablation for the embedding dimension of the projection.
LEAP is stable wrt.\  the dimension, performing consistently well. 
    }\resizebox{\linewidth}{!}{\begin{tabular}{ll SSS}
			\toprule
			{\small\sc Emb.}         & {\sc\small PE} & {\small\sc Letter-H} & {\sc\small Letter-M} & {\sc\small Letter-L} \\
			\midrule
			\multirow[c]{4}{*}{2}  & {LaPE}           & 66.31 \pm 1.2             & 94.71 \pm 1.29            & 76.76 \pm 1.95       \\
			                       & {RWPE}           & 73.82 \pm 2.0             & 94.62 \pm 0.51            & 83.47 \pm 1.68       \\
			                       & {\leaprd}        & \tb 81.16 \pm 1.66        & \tb 96.27 \pm 1.47        & \tsb 86.76 \pm 1.86       \\
			                       & {\leapld}        & \tsb 78.53 \pm 3.3        & \tsb 95.56 \pm 0.31       & \tb 87.38 \pm 0.99       \\
			\midrule
			\multirow[c]{4}{*}{5}  & {LaPE}           & 64.44 \pm 3.54            & 94.22 \pm 0.61            & 82.09 \pm 1.64       \\
			                       & {RWPE}           & 75.64 \pm 1.18            & 94.67 \pm 0.96            & 85.82 \pm 0.76       \\
			                       & {\leaprd}        & \tsb 79.78 \pm 0.57       & \tb  96.98 \pm 0.25       & \tsb 86.76 \pm 1.11       \\
			                       & {\leapld}        & \tb 80.4 \pm 1.41         & \tsb 96.44 \pm 0.59       & \tb 88.36 \pm 0.64       \\
			\midrule
			\multirow[c]{4}{*}{10} & {LaPE}           & 65.02 \pm 1.58            & 91.11 \pm 2.23            & 76.93 \pm 2.52       \\
			                       & {RWPE}           & 79.24 \pm 1.43            & 94.67 \pm 0.97            & 84.53 \pm 1.32       \\
			                       & {\leaprd}        & \tsb 80.13 \pm 2.04       & \tsb 96.68 \pm 0.78       & \tsb 86.59 \pm 2.01       \\
			                       & {\leapld}        & \tb 80.68 \pm 2.42        & \tb 96.99 \pm 1.05        & \tb 87.24 \pm 2.31       \\
			\midrule
			\multirow[c]{4}{*}{20} & {LaPE}           & 64.8 \pm 2.49             & 93.24 \pm 0.96            & 77.64 \pm 2.12       \\
			                       & {RWPE}           & 76.76 \pm 1.36            & 95.38 \pm 1.12            & 86.49 \pm 1.21       \\
			                       & {\leaprd}        & \tb 80.84 \pm 1.89        & \tsb 95.56 \pm 0.65       & \tb 87.42 \pm 0.75       \\
			                       & {\leapld}        & \tsb 79.91 \pm 1.53       & \tb 96.4 \pm 0.79         & \tsb 86.89 \pm 1.56       \\
			\bottomrule
		\end{tabular}
	}
	\label{tab:Ablation Embedding Dimension}
\end{wraptable}
 \paragraph{Projection strategies.}
We repeated all experiments using the five proposed LEAP projection strategies such
that each projection strategy has approximately similar small parameter budgets, comprising 1K--5K
parameters.
\cref{tab:LEAP All Projections} reports the results; and we find \emph{no} single
projection consistently outperformed the others, showing the best projection to be
dependent on the dataset--architecture combination.
However, a remarkable fact is that learnable directions did \emph{on average} outperform
the fixed set of directions, underpinning the benefits of learnable directions as compared
to using them as static features.

\paragraph{Locality parameter sensitivity.}
We also study the effect of the locality parameter in LEAP by repeating the
experiments on the \emph{Letter} datasets, originally performed with $1$-hop
neighborhoods, using instead $2$-hop neighborhoods and the concatenation of
LEAP embeddings from $1$- and $2$-hop neighborhoods.\footnote{In the concatenation setting, each embedding is computed with half the target
	dimension so that the final representation matches the dimension of the other
	approaches.
}
For this ablation, we use the NoMP model so that the models can only
access structural information through the PE, and we use  \emph{attention with PE} as the LEAP projection strategy.
\cref{tab:Ablation Hops} shows that the $1$-hop neighborhood yields the best performance across all
datasets, followed by the concatenated $1$- and $2$-hop version, respectively.

\paragraph{Impact of PE dimension.}
To better understand the effect of increasing the embedding dimension for the projection
strategies, we vary the size of the embedding dimension on the \emph{Letter} datasets.
The original experiment was ran with PE dimension $10$ for both LEAP and the baselines,
and we now repeat it with the embedding dimension set to $\{2,5,10,20\}$, respectively.
As before, we fix the architecture to NoMP so that models access structural
information only through the PE, and use \emph{attention with PE} as the
projection strategy for LEAP.
The results in \cref{tab:Ablation Embedding Dimension} show that
across \emph{all} evaluated PE dimensions and datasets, LEAP outperforms both
RWPE and LaPE.
\begin{table}
	\centering
	\sisetup{
		detect-all              = true,
		table-format            = 2.1(1.1),
		detect-mode             = true,
		separate-uncertainty    = true,
		retain-zero-uncertainty = true,
		mode                    = text,
	}\caption{Accuracy results for different PE strategies when using a variety of GNN architectures for multiple datasets from the TU Dataset benchmark.
        We evaluate the GCN, GAT and GIN architectures with and without LEAP.
        Additionally we evaluate LEAP on a non-message passing architecture~(NoMP).  
		Best results are \best{green}, second best are \secondbest{green}, and worst are \worst{red}.
		For every dataset, our approach achieves the best and second best results.}\resizebox{\linewidth}{!}{\begin{tabular}{@{}ll SSS SSSS @{}}
			\toprule
			{\small\sc Model}                      & {\small\sc PE}      & {\small\sc COX2}  & {\small\sc BZR}   & {\small\sc DHFR}  & {\small\sc Letter-H} & {\small\sc Letter-M} & {\small\sc Letter-L} & {\small\sc Fingerprint} \\
			\midrule
			\multirow[c]{5}{*}{\small\sc GCN}      & {\small No PE    }  & \tw77.9 \pm 1.0   & 81.9 \pm 3.3      & 71.6 \pm 1.4      & \tw 41.6\pm4.1       & \tw 63.5 \pm 2.0     & \tw 80.4 \pm 1.0     & 48.8 \pm 1.4            \\
			                                       & {\small RWPE     }  & 78.4 \pm 0.5      & \tw 79.5 \pm 2.2  & 73.0 \pm 2.4      & 60.9\pm 1.7          & 68.9 \pm 2.7         & 83.2 \pm 1.4         & 49.4 \pm 0.6            \\
			                                       & {\small LaPE     }  & 78.4 \pm 0.9      & 80.3 \pm 1.2      & \tw70.4 \pm 2.8   & 55.3\pm 2.6          & 75.8 \pm 2.6         & 89.2  \pm 1.2        & \tw 48.1 \pm 1.8        \\
			                                       & {\small \leaprd }   & \tsb 79.2 \pm 0.6 & \tsb82.5 \pm 2.4  & \tsb74.1 \pm 5.2  & \tsb 72.2 \pm 3.3    & \tsb 82.6 \pm 1.4    & \tsb 95.8 \pm 1.1    & \tb 55.6 \pm 1.1        \\
			                                       & {\small \leapld }   & \tb 79.4 \pm 1.0  & \tb82.5 \pm 1.6   & \tb77.6 \pm 2.8   & \tb 74.2 \pm  1.5    & \tb 83.6 \pm 1.3     & \tb 96.0 \pm 0.9     & \tsb 55.1 \pm 1.2       \\
			\midrule
			\multirow[c]{5}{*}{\small\sc GAT}      & {\small No PE    }  & 78.2 \pm 0.6      & 80.5 \pm 2.0      & 73.7 \pm 1.8      & \tw 41.9 \pm 3.2     & \tw 58.4 \pm 3.7     & \tw 89.4 \pm 0.7     & 50.5 \pm 0.6            \\
			                                       & {\small RWPE     }  & 79.0 \pm 1.4      & \tw78.3 \pm 1.1   & 70.9 \pm 2.4      & 63.0 \pm 3.0         & 69.0 \pm 1.8         & 90.8 \pm 1.5         & 50.4 \pm 0.8            \\
			                                       & {\small LaPE     }  & \tw 77.9 \pm 1.0  & 80.3 \pm 1.2      & \tw70.4 \pm 2.7   & 54.7 \pm 5.3         & 75.2 \pm 2.3         & 89.6 \pm 1.5         & \tw 48.9 \pm 1.0        \\
			                                       & {\small \leaprd }   & \tsb 79.2 \pm 1.6 & \tsb82.0 \pm 3.2  & \tsb75.7 \pm 3.0  & \tsb 70.2 \pm 2.2    & \tb 83.2 \pm 1.1     & \tb 95.8 \pm 0.8     & \tb 55.1 \pm 0.6        \\
			                                       & {\small \leapld }   & \tb 80.1 \pm 2.2  & \tb83.7 \pm 2.9   & \tb76.5 \pm 3.8   & \tb 73.5 \pm 2.1     & \tsb 82.4 \pm 1.6    & \tsb 95.2 \pm 0.9    & \tsb 54.9 \pm 0.7       \\
			\midrule
			\multirow[c]{5}{*}{\small\sc GIN}      & {\small  No PE    } & \tw 78.2 \pm 0.5  & 79.5 \pm 1.4      & 69.3 \pm 4.8      & \tw 47.7 \pm 0.8     & 65.0 \pm 3.9         & 82.7 \pm 1.5         & \tw 48.4 \pm 1.4        \\
			                                       & {\small  RWPE     } & 78.6 \pm 1.6      & \tw 79.3 \pm 1.0  & 72.0 \pm 4.9      & 54.4 \pm 2.3         & \tw 64.9 \pm 3.7     & \tw 81.6 \pm 2.4     & 50.0 \pm 1.6            \\
			                                       & {\small  LaPE     } & \tw 78.2 \pm 0.5  & 79.5 \pm 1.7      & \tw 61.0 \pm 0.1  & 55.0 \pm 3.5         & 75.2 \pm 3.5         & 84.4 \pm 3.5         & 49.8 \pm 1.9            \\
			                                       & {\small \leaprd }   & \tsb 79.0 \pm 0.9 & \tb 81.2 \pm 1.4  & \tsb 73.9 \pm 4.1 & \tsb 60.2 \pm 4.8    & \tsb 76.3 \pm 2.0    & \tsb 93.3 \pm 1.1    & \tsb 54.4 \pm 1.1       \\
			                                       & {\small \leapld }   & \tb 79.6 \pm 1.6  & \tsb 81.0 \pm 1.4 & \tb 76.2 \pm 3.2  & \tb 62.7 \pm 3.4     & \tb 77.6 \pm 1.7     & \tb 94.0 \pm 1.9     & \tb 55.3 \pm 1.4        \\
			\midrule
			\multirow[c]{5}{*}{\small\sc NoMP}     & {\small No PE    }  & 77.9 \pm 0.8      & \tw79.8 \pm 2.6   & \tw70.1 \pm 3.4   & \tw 63.4 \pm 1.0     & \tw 57.8 \pm 0.9     & \tw 89.7 \pm 1.3     & 50.7 \pm 0.5            \\
			                                       & {\small RWPE     }  & \tw77.7 \pm 1.3   & 80.9 \pm 1.7      & 73.3 \pm 1.5      & 79.2 \pm 1.4         & 84.5 \pm 1.3         & 94.7 \pm 1.0         & 51.3 \pm 0.7            \\
			                                       & {\small LaPE     }  & 77.7 \pm 1.0      & 81.2 \pm 3.2      & 70.5 \pm 3.5      & 65.0 \pm 1.6         & 76.9 \pm 2.5         & 91.1 \pm 2.2         & \tw 50.5 \pm 1.2        \\
			                                       & {\small \leaprd }   & \tb 79.0 \pm 0.6  & \tsb83.2 \pm 1.7  & \tsb74.3 \pm 6.1  & \tsb 81.3 \pm 1.9    & \tsb 88.0 \pm 1.9    & \tsb 97.2 \pm 0.3    & \tsb 55.7 \pm 1.1       \\
			                                       & {\small \leapld }   & \tsb78.6 \pm 0.8  & \tb84.7 \pm 2.7   & \tb75.7 \pm 2.7   & \tb 81.6 \pm 1.9     & \tb 88.5 \pm 2.5     & \tb 98.0 \pm 0.4     & \tb 56.3 \pm 1.4        \\
\bottomrule
		\end{tabular}
	}
\label{tab:LEAP Best Projection}
\end{table}

\paragraph{DECT hyperparameters and comparison.}

We assessed LEAP’s sensitivity to the DECT hyperparameters by varying the number of directions in $\{2,4,8,16,32\}$ and smoothing parameter in $\{2,4,8,16,32,64,128\}$. LEAP remained robust, outperforming baselines across all settings; see \cref{fig:DECT Hyperparameter ablation}.
Finally, \cref{tab:LEAP DECT Comparison} summarizes the comparison of LEAP with DECT for graph classification tasks.
LEAP outperforms two variants of the DECT~(with different parameter budgets) on most datasets, which further underscores the utility of learnable directions.

\section{Conclusion and future work}

We presented LEAP, a new \emph{learnable local structural positional encoding} for graphs based on the \lect. To the best of our knowledge, this is the \emph{first} approach to integrate the \lect{} into deep learning architectures in an end-to-end trainable fashion.

Our experiments show that LEAP consistently outperforms established baselines across multiple architectures and datasets, with learned directions further improving performance in most tasks, thereby highlighting the benefits of making this step trainable. Additionally, we introduced a synthetic task in which our approach achieved perfect accuracy, demonstrating its ability to capture topological information independent of node features, which the evaluated MPNNs (GCN/GAT) alone failed to recover.
Taken together, these results highlight the potential of \lect{} encodings for \emph{topological deep learning}~\citep{papamarkou2024position} and graph representation learning tasks.
LEAP is particularly well-suited to provide local structural information to architectures that rely on global attention mechanisms, where graph structure is not directly modeled and multiple PEs are combined to capture complementary notions of graph position.

\paragraph{Limitations.} While LEAP provides a learnable way to capture local structural information it has some limitations. First, it is not a purely structural PE, as it requires \emph{node features} to compute the ECTs. However, these features can be \emph{learned}, and in the synthetic dataset, our approach succeeded even though the features were irrelevant to the prediction targets.
Second, LEAP relies on a differentiable approximation of the discretized ECT applied to normalized $m$-hop subgraphs, which are not necessarily geometric, so the theoretical guarantees of the \emph{exact} ECT~(e.g., injectivity) may not fully carry over; we expect this to be interesting for future work. Finally, unlike other graph PEs, such as LaPE or RWPE, where the only hyperparameter is the embedding dimension, LEAP introduces several hyperparameters~(among others, a smoothing parameter of the ECT approximation, the number of directions, and the number of discretization steps). In practice, however, we fixed these across datasets and architectures, nevertheless observing consistently strong performance. Our ablation studies further serve to demonstrate the robustness to these choices.

\paragraph{Future work.} We envision several directions for future research.
First, drawing on prior work~\citep{lect}, we aim to formalize the theoretical expressivity of LEAP, noting that theoretical expressivity and empirical performance are often not correlated.
Combining LEAP with positional encodings that capture complementary aspects of graph structure and embedding it within more sophisticated architectures may further improve performance and expressivity. 
Another promising line of work is to make the ECT step fully differentiable. 
Instead of discretizing along a fixed grid, treating thresholds as trainable parameters would allow the model to 
focus on informative regions and optimize their positions jointly with the other parameters.
Using learned features, we also plan on assessing the performance of LEAP on non-attributed graph datasets, i.e., datasets that are fully structural.
Finally, since the ECT can be applied to higher-order datasets~\citep{Ballester25a} comprising, for instance, simplicial complexes or cell complexes~\citep{Hoppe25a}, 
we believe that LEAP could be extended to this modality, thus serving as a generalizable 
addition to the topological deep learning toolbox.

\subsubsection*{Reproducibility Statement}
Our code can be accessed at \url{https://github.com/aidos-lab/LEAP}.
All experiments used a fixed seed; the full configurations can be found in 
the \texttt{experiments} folder.

\subsubsection*{Acknowledgments}

\textit{In loving memory of Teresa Paz Camps.}

First and foremost, facing a turbulent decision-making process, we want to thank the reviewers and the AC for believing in the merits of our work. We are particularly grateful for the support by reviewer \texttt{f41T}, who continued to champion our work throughout the revision process.
This work has received funding from the Swiss State Secretariat for
Education, Research, and Innovation~(SERI).
B.R.\ wishes to dedicate this paper to his daughter Aur\'elie.

\clearpage
\bibliographystyle{iclr2026_conference}
\bibliography{iclr2026_conference}

@article{Rieck25,
	archiveprefix = {arXiv},
	author = {Bastian Rieck},
	eprint = {2410.17760},
	journal = {Notices of the American Mathematical Society},
	number = {7},
	pages = {719--727},
	primaryclass = {cs.LG},
	title = {Topology meets Machine Learning: An Introduction using the
	         {E}uler Characteristic Transform},
	volume = {72},
	year = {2025},
}

@article{bench-gnns,
	title = {Benchmarking Graph Neural Networks},
	author = {Dwivedi, Vijay Prakash and Joshi, Chaitanya K and Luu, Anh
	          Tuan and Laurent, Thomas and Bengio, Yoshua and Bresson, Xavier
	          },
	journal = {Journal of Machine Learning Research},
	volume = {24},
	number = {43},
	pages = {1--48},
	year = {2023},
}

@article{bench-pe,
	title = {Benchmarking positional encodings for GNNs and graph transformers},
	author = {Gr{\"o}tschla, Florian and Xie, Jiaqing and Wattenhofer, Roger},
	journal = {arXiv preprint arXiv:2411.12732},
	archiveprefix = {arXiv},
	year = {2024},
}

@inproceedings{dect,
	title = {Differentiable {E}uler Characteristic Transforms for Shape
	         Classification},
	author = {Ernst R{\"o}ell and Bastian Rieck},
	year = 2024,
	booktitle = {International Conference on Learning Representations},
	eprint = {2310.07630},
	archiveprefix = {arXiv},
	primaryclass = {cs.LG},
	repository = {https://github.com/aidos-lab/dect-evaluation},
}

@article{ect-invitation,
	title = {An invitation to the {E}uler characteristic transform},
	author = {Munch, Elizabeth},
	journal = {The American Mathematical Monthly},
	volume = {132},
	number = {1},
	pages = {15--25},
	year = {2025},
	publisher = {Taylor \& Francis},
}

@inproceedings{gat-pe,
	title = {Graph Attention Networks with Positional Embeddings},
	author = {Ma, Liheng and Rabbany, Reihaneh and Romero-Soriano, Adriana},
	year = 2021,
	booktitle = {Advances in Knowledge Discovery and Data Mining},
	publisher = {Springer},
	address = {Cham, Switzerland},
	pages = {514--527},
	editor = {Karlapalem, Kamal and Cheng, Hong and Ramakrishnan, Naren and
	          Agrawal, R. K. and Reddy, P. Krishna and Srivastava, Jaideep
	          and Chakraborty, Tanmoy},
}

@article{gen-lape,
	title = {Generalized {L}aplacian positional encoding for graph
	         representation learning},
	author = {Maskey, Sohir and Parviz, Ali and Thiessen, Maximilian and St{\"a}rk , Hannes and Sadikaj, Ylli and Maron, Haggai},
	journal = {arXiv preprint arXiv:2210.15956},
	archiveprefix = {arXiv},
	year = {2022},
}

@inproceedings{learn-pe,
	title = {Graph Neural Networks with Learnable Structural and Positional
	         Representations},
	author = {Vijay Prakash Dwivedi and Anh Tuan Luu and Thomas Laurent and
	          Yoshua Bengio and Xavier Bresson},
	booktitle = {International Conference on Learning Representations},
	year = {2022},
}

@inproceedings{gps,
  title         = {Recipe for a General, Powerful, Scalable Graph Transformer},
  author        = {Ramp\'{a}\v{s}ek, Ladislav and Galkin, Michael and Dwivedi, Vijay Prakash and Luu, Anh Tuan and Wolf, Guy and Beaini, Dominique},
  year          = 2022,
  booktitle     = {Advances in Neural Information Processing Systems},
  publisher     = {Curran Associates, Inc.},
  volume        = 35,
  pages         = {14501--14515},
  editor        = {S. Koyejo and S. Mohamed and A. Agarwal and D. Belgrave and K. Cho and A. Oh}
}

@article{inverse-ect,
	title = {Persistent homology and {E}uler integral transforms},
	author = {Ghrist, Robert and Levanger, Rachel and Mai, Huy},
	journal = {Journal of Applied and Computational Topology},
	volume = {2},
	number = {1},
	pages = {55--60},
	year = {2018},
	publisher = {Springer},
}

@inproceedings{point-transform,
  title         = {Point Cloud Synthesis Using Inner Product Transforms},
  author        = {Ernst R{\"o}ell and Bastian Rieck},
  year          = 2025,
  booktitle     = {Advances in Neural Information Processing Systems},
  publisher     = {Curran Associates, Inc.},
  volume        = 38,
  pages         = {14551--14583},
  archiveprefix = {arXiv},
  author+an     = {2=highlight},
  editor        = {D. Belgrave and C. Zhang and H. Lin and R. Pascanu and P. Koniusz and M. Ghassemi and N. Chen},
  eprint        = {2410.18987},
  primaryclass  = {cs.CV},
}

@inproceedings{k-forms,
	archiveprefix = {arXiv},
	author = {Kelly Maggs and Celia Hacker and Bastian Rieck},
	author+an = {3=highlight},
	booktitle = {International Conference on Learning Representations},
	eprint = {2312.08515},
	primaryclass = {cs.LG},
	repository = {https://github.com/aidos-lab/neural-k-forms},
	title = {Simplicial Representation Learning with Neural $k$-Forms},
	year = {2024},
}

@inproceedings{lect,
  title         = {Diss-l-{ECT}: Dissecting Graph Data with Local {E}uler Characteristic Transforms},
  author        = {von Rohrscheidt, Julius and Rieck, Bastian},
  year          = 2025,
  booktitle     = {Proceedings of the 42nd International Conference on Machine Learning},
  publisher     = {PMLR},
  series        = {Proceedings of Machine Learning Research},
  volume        = 267,
  pages         = {61790--61809},
  editor        = {Singh, Aarti and Fazel, Maryam and Hsu, Daniel and Lacoste-Julien, Simon and Berkenkamp, Felix and Maharaj, Tegan and Wagstaff, Kiri and Zhu, Jerry},
}

@inproceedings{gcn,
	title = {Semi-Supervised Classification with Graph Convolutional
	         Networks},
	author = {Thomas N. Kipf and Max Welling},
	booktitle = {International Conference on Learning Representations},
	year = {2017},
}

@inproceedings{attention,
  title         = {Attention is All you Need},
  author        = {Vaswani, Ashish and Shazeer, Noam and Parmar, Niki and Uszkoreit, Jakob and Jones, Llion and Gomez, Aidan N and Kaiser, \L ukasz and Polosukhin, Illia},
  year          = 2017,
  booktitle     = {Advances in Neural Information Processing Systems},
  publisher     = {Curran Associates, Inc.},
  volume        = 30,
  pages         = {5998--6008},
  editor        = {I. Guyon and U. Von Luxburg and S. Bengio and H. Wallach and R. Fergus and S. Vishwanathan and R. Garnett}
}

@article{moleculenet,
	title = {{MoleculeNet}: {A} benchmark for molecular machine learning},
	author = {Wu, Zhenqin and Ramsundar, Bharath and Feinberg, Evan N and
	          Gomes, Joseph and Geniesse, Caleb and Pappu, Aneesh S and
	          Leswing, Karl and Pande, Vijay},
	journal = {Chemical Science},
	volume = {9},
	number = {2},
	pages = {513--530},
	year = {2018},
	publisher = {Royal Society of Chemistry},
}

@article{tu-datasets,
	title = {{TUDataset}: {A} collection of benchmark datasets for learning
	         with graphs},
	author = {Morris, Christopher and Kriege, Nils M and Bause, Franka and
	          Kersting, Kristian and Mutzel, Petra and Neumann, Marion},
	journal = {arXiv preprint arXiv:2007.08663},
	archiveprefix = {arXiv},
	year = {2020},
}

@article{alchemy,
	title = {Alchemy: {A} quantum chemistry dataset for benchmarking {AI}
	         models},
	author = {Chen, Guangyong and Chen, Pengfei and Hsieh, Chang-Yu and Lee,
	          Chee-Kong and Liao, Benben and Liao, Renjie and Liu, Weiwen and
	          Qiu, Jiezhong and Sun, Qiming and Tang, Jie and others},
	journal = {arXiv preprint arXiv:1906.09427},
	archiveprefix = {arXiv},
	year = {2019},
}

@article{mlp,
	title = {Learning representations by back-propagating errors},
	author = {Rumelhart, David E and Hinton, Geoffrey E and Williams, Ronald
	          J},
	journal = {Nature},
	volume = {323},
	number = {6088},
	pages = {533--536},
	year = {1986},
	publisher = {Nature Publishing Group UK London},
}

@inproceedings{mpnn,
  title         = {Neural Message Passing for Quantum Chemistry},
  author        = {Justin Gilmer and Samuel S. Schoenholz and Patrick F. Riley and Oriol Vinyals and George E. Dahl},
  year          = 2017,
  booktitle     = {Proceedings of the 34th International Conference on Machine Learning},
  publisher     = {PMLR},
  series        = {Proceedings of Machine Learning Research},
  volume        = 70,
  pages         = {1263--1272},
  editor        = {Precup, Doina and Teh, Yee Whye},
}

@article{ect,
	title = {Persistent homology transform for modeling shapes and surfaces},
	author = {Turner, Katharine and Mukherjee, Sayan and Boyer, Doug M},
	journal = {Information and Inference: A Journal of the IMA},
	volume = {3},
	number = {4},
	pages = {310--344},
	year = {2014},
	publisher = {Oxford University Press},
}

@article{oversmoothing1,
	title = {A survey on oversmoothing in graph neural networks},
	author = {Rusch, T Konstantin and Bronstein, Michael M and Mishra,
	          Siddhartha},
	journal = {arXiv preprint arXiv:2303.10993},
	archiveprefix = {arXiv},
	year = {2023},
}

@inproceedings{oversmoothing2,
	title = {A comprehensive review of the oversmoothing in graph neural networks},
	author = {Zhang, Xu and Xu, Yonghui and He, Wei and Guo, Wei and Cui,
	          Lizhen},
	booktitle = {CCF Conference on Computer Supported Cooperative Work and
	             Social Computing},
	pages = {451--465},
	year = {2023},
	organization = {Springer},
}

@inproceedings{oversquashing,
  title         = {On Over-Squashing in Message Passing Neural Networks: The Impact of Width, Depth, and Topology},
  author        = {Di Giovanni, Francesco and Giusti, Lorenzo and Barbero, Federico and Luise, Giulia and Lio, Pietro and Bronstein, Michael M.},
  year          = 2023,
  booktitle     = {Proceedings of the 40th International Conference on Machine Learning},
  publisher     = {PMLR},
  series        = {Proceedings of Machine Learning Research},
  volume        = 202,
  pages         = {7865--7885},
  editor        = {Krause, Andreas and Brunskill, Emma and Cho, Kyunghyun and Engelhardt, Barbara and Sabato, Sivan and Scarlett, Jonathan},
}

@inproceedings{counting,
  title         = {Can Graph Neural Networks Count Substructures?},
  author        = {Chen, Zhengdao and Chen, Lei and Villar, Soledad and Bruna, Joan},
  year          = 2020,
  booktitle     = {Advances in Neural Information Processing Systems},
  publisher     = {Curran Associates, Inc.},
  volume        = 33,
  pages         = {10383--10395},
  editor        = {H. Larochelle and M. Ranzato and R. Hadsell and M.F. Balcan and H. Lin}
}

@inproceedings{howpowerful,
	title = {How Powerful are Graph Neural Networks?},
	author = {Keyulu Xu and Weihua Hu and Jure Leskovec and Stefanie Jegelka},
	booktitle = {International Conference on Learning Representations},
	year = {2019},
}

@inproceedings{virtualnode1,
	title = {On the Connection Between {MPNN} and Graph Transformer},
	author = {Cai, Chen and Hy, Truong Son and Yu, Rose and Wang, Yusu},
	booktitle = {Proceedings of the 40th International Conference on Machine Learning},
	pages = {3408--3430},
	year = {2023},
	editor = {Krause, Andreas and Brunskill, Emma and Cho, Kyunghyun and
	          Engelhardt, Barbara and Sabato, Sivan and Scarlett, Jonathan},
	volume = {202},
	series = {Proceedings of Machine Learning Research},
	publisher = {PMLR},
}

@article{ectdirections,
	title = {How many directions determine a shape and other sufficiency
	         results for two topological transforms},
	author = {Curry, Justin and Mukherjee, Sayan and Turner, Katharine},
	journal = {Transactions of the American Mathematical Society, Series B},
	volume = {9},
	number = {32},
	pages = {1006--1043},
	year = {2022},
}

@inproceedings{deepsets,
	title = {Deep Sets},
	author = {Zaheer, Manzil and Kottur, Satwik and Ravanbakhsh, Siamak and
	          Poczos, Barnabas and Salakhutdinov, Russ R and Smola, Alexander
	          J},
	year = 2017,
	booktitle = {Advances in Neural Information Processing Systems},
	publisher = {Curran Associates, Inc.},
	volume = 30,
	pages = {3391--3401},
	editor = {I. Guyon and U. Von Luxburg and S. Bengio and H. Wallach and
	          R. Fergus and S. Vishwanathan and R. Garnett},
}

@inproceedings{tope,
  title         = {Positional Encoding meets Persistent Homology on Graphs},
  author        = {Verma, Yogesh and Souza, Amauri H and Garg, Vikas K},
  year          = 2025,
  booktitle     = {Proceedings of the 42nd International Conference on Machine Learning},
  publisher     = {PMLR},
  series        = {Proceedings of Machine Learning Research},
  volume        = 267,
  pages         = {61323--61343},
  editor        = {Singh, Aarti and Fazel, Maryam and Hsu, Daniel and Lacoste-Julien, Simon and Berkenkamp, Felix and Maharaj, Tegan and Wagstaff, Kiri and Zhu, Jerry},
}

@article{Crawford20a,
	title = {Predicting Clinical Outcomes in Glioblastoma: {A}n Application
	         of Topological and Functional Data Analysis},
	author = {Lorin Crawford and Anthea Monod and Andrew X. Chen and Sayan
	          Mukherjee and Raúl Rabadán},
	year = 2020,
	journal = {Journal of the American Statistical Association},
	publisher = {Taylor {\&} Francis},
	volume = 115,
	number = 531,
	pages = {1139--1150},
}

@article{cisewskikehe2023weighted,
	title = {The Weighted {E}uler Characteristic Transform for Image Shape
	         Classification},
	author = {Jessi Cisewski-Kehe and Brittany Terese Fasy and Dhanush
	          Giriyan and Eli Quist},
	year = {2023},
	journal = {arXiv preprint arXiv:2307.13940},
	archiveprefix = {arXiv},
	primaryClass = {cs.CG},
}

@article{amezquita2022measuring,
	title = {Measuring hidden phenotype: {Q}uantifying the shape of barley
	         seeds using the {E}uler characteristic transform},
	author = {Amézquita, Erik J and Quigley, Michelle Y and Ophelders, Tim
	          and Landis, Jacob B and Koenig, Daniel and Munch, Elizabeth and
	          Chitwood, Daniel H},
	year = 2021,
	month = 12,
	journal = {in silico Plants},
	volume = 4,
	number = 1,
	pages = {diab033},
}

@inproceedings{jiang2020weighted,
	title = {The Weighted {E}uler Curve Transform for Shape and Image
	         Analysis},
	author = {Q. Jiang and S. Kurtek and T. Needham},
	year = 2020,
	booktitle = {Proceedings of the IEEE/CVF Conference on Computer Vision
	             and Pattern Recognition Workshops~(CVPRW)},
	pages = {3685--3694},
}

@inproceedings{Horn22a,
	author = {Horn, Max and {De Brouwer}, Edward and Moor, Michael and
	          Moreau, Yves and Rieck, Bastian and Borgwardt, Karsten},
	title = {Topological Graph Neural Networks},
	year = {2022},
	booktitle = {International Conference on Learning Representations},
}

@inproceedings{gat,
	title = {Graph Attention Networks},
	author = {Petar Veličković and Guillem Cucurull and Arantxa Casanova and
	          Adriana Romero and Pietro Liò and Yoshua Bengio},
	year = 2018,
	booktitle = {International Conference on Learning Representations},
}

@inproceedings{signNet,
	title = {Sign and Basis Invariant Networks for Spectral Graph
	         Representation Learning},
	author = {Derek Lim and Joshua David Robinson and Lingxiao Zhao and Tess
	          Smidt and Suvrit Sra and Haggai Maron and Stefanie Jegelka},
	booktitle = {International Conference on Learning Representations},
	year = {2023},
}

@inproceedings{Kreuzer21a,
 author = {Kreuzer, Devin and Beaini, Dominique and Hamilton, Will and L\'{e}tourneau, Vincent and Tossou, Prudencio},
 booktitle = {Advances in Neural Information Processing Systems},
 editor = {M. Ranzato and A. Beygelzimer and Y. Dauphin and P.S. Liang and J. Wortman Vaughan},
 pages = {21618--21629},
 publisher = {Curran Associates, Inc.},
 title = {Rethinking Graph Transformers with Spectral Attention},
 volume = {34},
 year = {2021}
}

@inproceedings{tnn,
	title = {Topological Neural Networks go Persistent, Equivariant, and
	         Continuous},
	author = {Verma, Yogesh and Souza, Amauri H and Garg, Vikas},
	year = 2024,
	booktitle = {Proceedings of the 41st International Conference on Machine Learning},
	publisher = {PMLR},
	series = {Proceedings of Machine Learning Research},
	volume = 235,
	pages = {49388--49407},
	editor = {Salakhutdinov, Ruslan and Kolter, Zico and Heller, Katherine
	          and Weller, Adrian and Oliver, Nuria and Scarlett, Jonathan and
	          Berkenkamp, Felix},
}

@inproceedings{papamarkou2024position,
	title = {Position: Topological Deep Learning is the New Frontier for Relational Learning},
	author = {Theodore Papamarkou and Tolga Birdal and Michael Bronstein and
	          Gunnar Carlsson and Justin Curry and Yue Gao and Mustafa Hajij
	          and Roland Kwitt and Pietro Liò and Paolo Di Lorenzo and
	          Vasileios Maroulas and Nina Miolane and Farzana Nasrin and
	          Karthikeyan Natesan Ramamurthy and Bastian Rieck and Simone
	          Scardapane and Michael T. Schaub and Petar Veličković and Bei
	          Wang and Yusu Wang and Guo-Wei Wei and Ghada Zamzmi},
	year = 2024,
	booktitle = {Proceedings of the 41st International Conference on Machine
	             Learning},
	publisher = {PMLR},
	series = {Proceedings of Machine Learning Research},
	pages = {39529--39555},
	volume = {235},
	editor = {Salakhutdinov, Ruslan and Kolter, Zico and Heller, Katherine
	          and Weller, Adrian and Oliver, Nuria and Scarlett, Jonathan and
	          Berkenkamp, Felix},
}

@article{Hoppe25a,
	title = {Don't be Afraid of Cell Complexes! {A}n Introduction from an
	         Applied Perspective},
	author = {Josef Hoppe and Vincent P. Grande and Michael T. Schaub},
	year = {2025},
	journal = {arXiv preprint arXiv:2506.09726},
	primaryclass = {eess.SP},
}

@inproceedings{Ballester25a,
	title = {{MANTRA}: The Manifold Triangulations Assemblage},
	author = {Rubén Ballester and Ernst Röell and Daniel Bīn Schmid and
	          Mathieu Alain and Sergio Escalera and Carles Casacuberta and
	          Bastian Rieck},
	year = 2025,
	eprint = {2410.02392},
	archiveprefix = {arXiv},
	primaryclass = {cs.LG},
	booktitle = {International Conference on Learning Representations},
}

@inproceedings{RomanEmpire,
	title = {A critical look at the evaluation of GNNs under heterophily:
	         Are we really making progress?},
	author = {Platonov, Oleg and Kuznedelev, Denis and Diskin, Michael and
	          Babenko, Artem and Prokhorenkova, Liudmila},
	year = 2023,
	eprint = {2302.11640},
	archiveprefix = {arXiv},
	primaryclass = {cs.LG},
	booktitle = {International Conference on Learning Representations},
}

\clearpage
\appendix
\newpage

\setcounter{figure}{0}
\setcounter{table}{0}
\renewcommand{\thefigure}{S.\arabic{figure}}
\renewcommand{\thetable}{S.\arabic{table}}

\crefalias{section}{appendix}

\makeatletter
\def\addcontentsline#1#2#3{\addtocontents{#1}{\protect\contentsline{#2}{#3}{\thepage}{\@currentHref}}}
\makeatother

\startcontents
\printcontents{}{1}{{\vskip5pt\hrule
    \large\textbf{Appendix~(Supplementary Materials)}\vskip3pt\hrule\vskip5pt}
}

\vskip15pt
\hrule
\vskip5pt
\clearpage

\section{Results for the Alchemy and HIV datasets}
\begin{figure}[h]
	\centering
	\begin{subfigure}[t]{1.0\textwidth}
		\begin{subfigure}{0.49\textwidth}
			\centering
			\includegraphics[width=\linewidth]{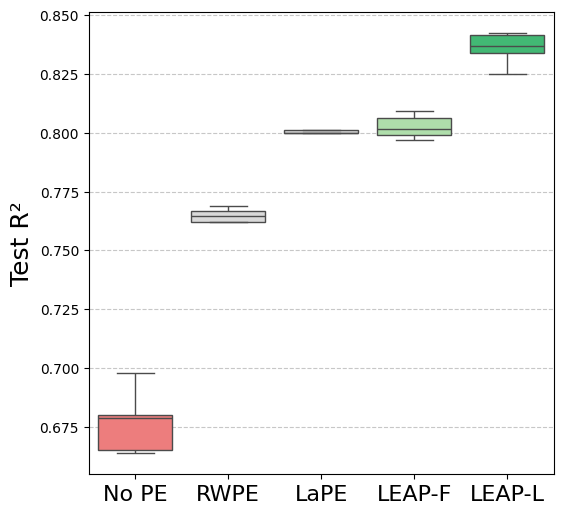}
			\caption*{GCN}
		\end{subfigure}\hfill
		\begin{subfigure}{0.49\textwidth}
			\centering
			\includegraphics[width=\linewidth]{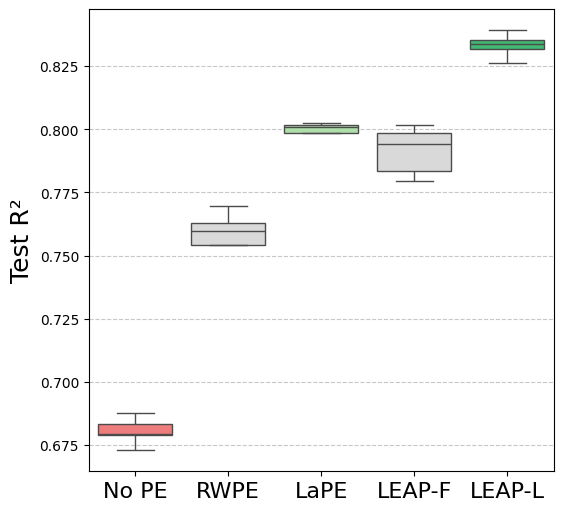}
			\caption*{GAT}
		\end{subfigure}
	\end{subfigure}

	\caption{$R^2$ results for different PE strategies on the \emph{Alchemy} dataset
		using the GCN and GAT architectures.
Best results in terms of mean $R^2$ are \best{green}, second best are \secondbest{green}, and worst are
		\worst{red}.
}\label{fig:Box Plot Alchemy Full}
\end{figure}

\begin{figure}[h]
	\centering
	\begin{subfigure}[t]{1.0\textwidth}
		\begin{subfigure}{0.49\textwidth}
			\centering
			\includegraphics[width=\linewidth]{images/HIV-REVIEW.png}
			\caption*{GCN}
		\end{subfigure}\hfill
		\begin{subfigure}{0.49\textwidth}
			\centering
			\includegraphics[width=\linewidth]{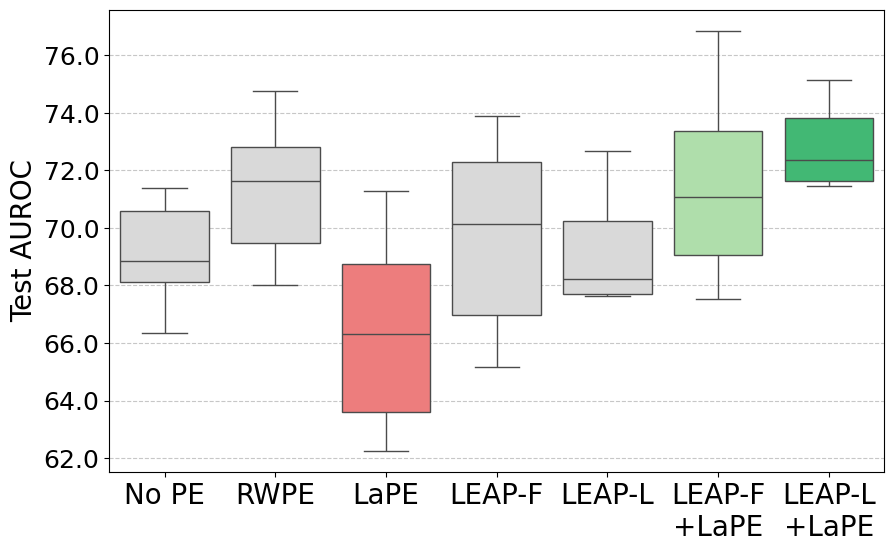}
			\caption*{GAT}
		\end{subfigure}
	\end{subfigure}
	\caption{AUROC results for different PE strategies on the \emph{HIV} dataset
		using the GCN and GAT architectures, respectively.
Best results in terms of mean AUROC are \best{green}, second best are \secondbest{green}, and worst are
		\worst{red}.
}\label{fig:Box Plot HIV}
\end{figure}
\clearpage
\section{Validation metrics}
\begin{figure}[h]
	\centering
	\begin{subfigure}{0.48\textwidth}
		\includegraphics[width=\linewidth]{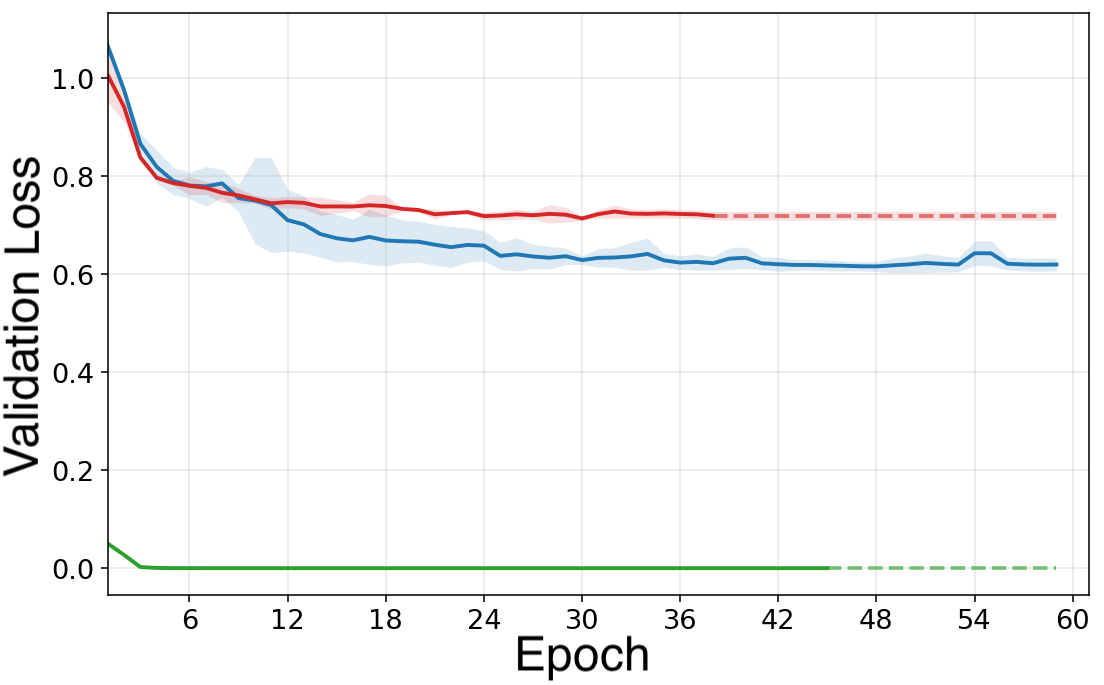}
		\caption*{Validation Loss}
		\label{fig:syn-1}
	\end{subfigure}\hfill
	\begin{subfigure}{0.48\textwidth}
		\includegraphics[width=\linewidth]{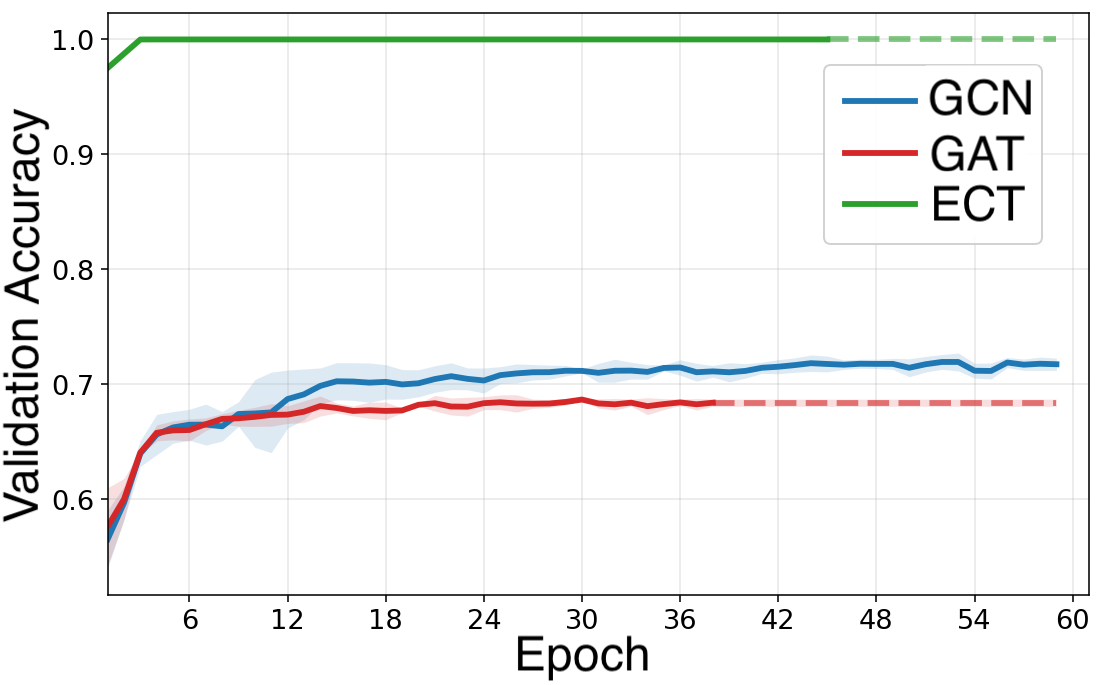}
		\caption*{Validation Accuracy}
		\label{fig:syn-2}
	\end{subfigure}
	\caption{Validation loss and accuracy per training epoch  for the synthetic dataset for the baseline GCN, GAT, and LEAP.
Our method achieves a perfect score in both metrics and convergence immediately.
The shadows indicate one standard deviation over 5 runs and the dashed line means that model training  finished earlier because of early stopping.
	}\label{fig:syn}
\end{figure}

\begin{figure}[h]
	\begin{subfigure}{0.48\textwidth}
		\includegraphics[width=0.9\linewidth]{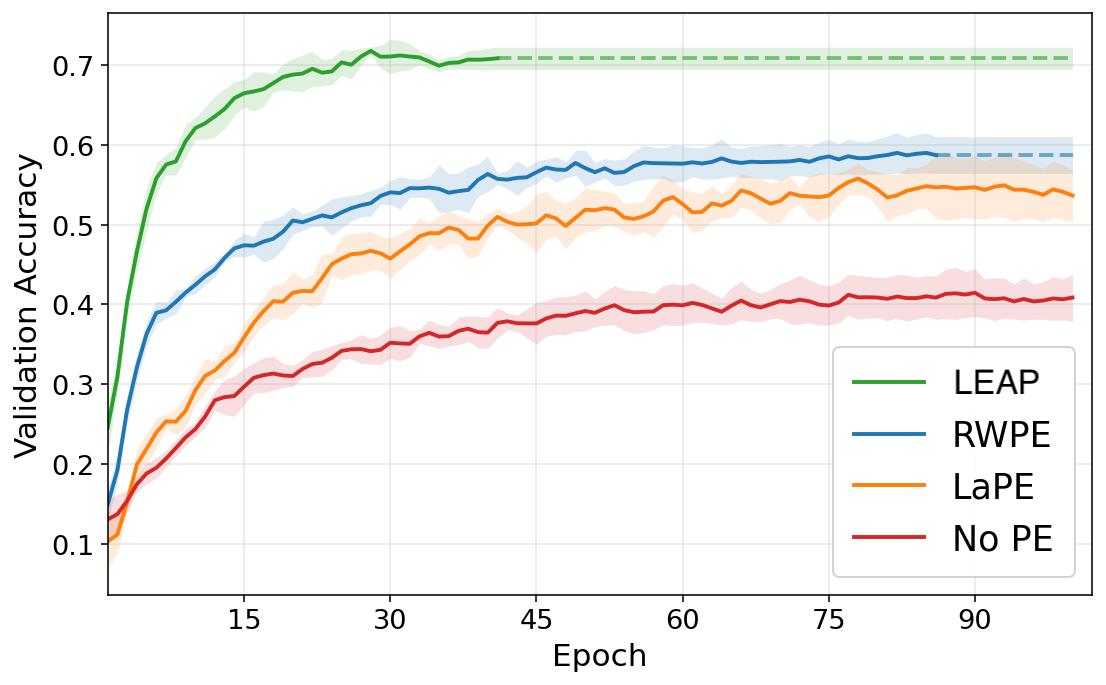}
		\caption*{Letter High}
		\label{fig:train-high}
	\end{subfigure}
	\hfill
	\begin{subfigure}{0.48\textwidth}
		\includegraphics[width=0.9\linewidth]{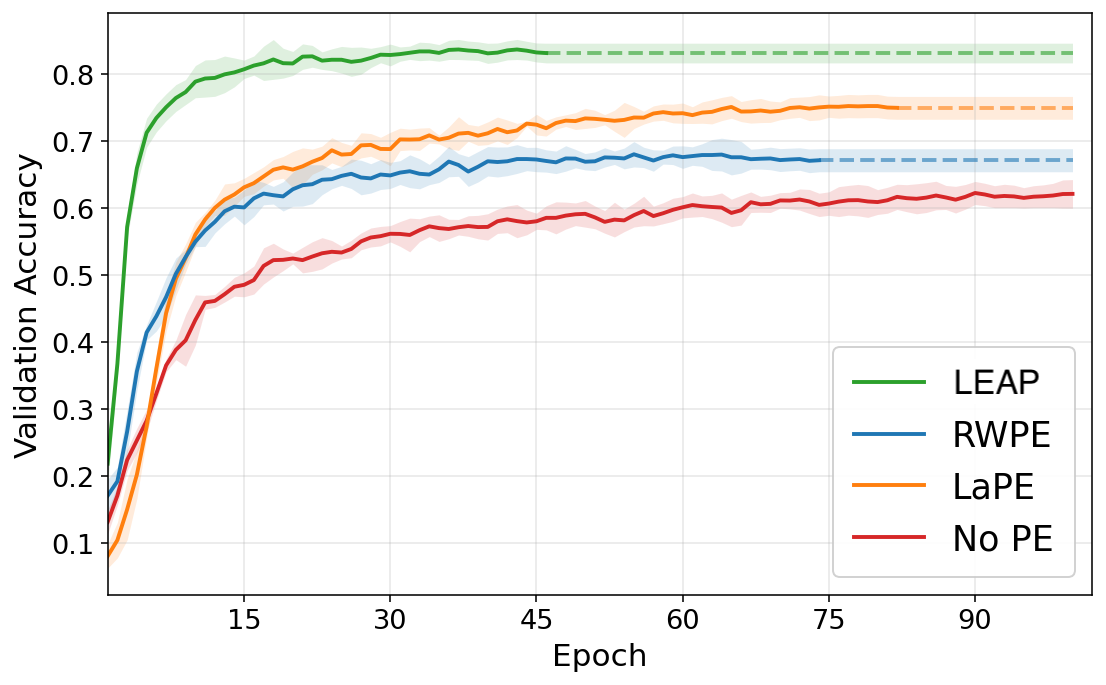}
		\caption*{Letter Medium}
		\label{fig:train-med}
	\end{subfigure}
	\caption{Validation accuracy per training epoch for the Letter High (left) and Letter Medium (right) datasets for different PE strategies using a GCN architecture.
Our method achieves the best results and converges faster.
The shadows around the curves indicate the standard deviation over $5$ runs and the dashed line means that training ended due to early stopping.
	}\end{figure}

\clearpage
\section{Additional experiments}
\begin{table}[h]
	\centering
	\sisetup{
		detect-all              = true,
		table-format            = 2.1(1.1),
		detect-mode             = true,
		separate-uncertainty    = true,
		retain-zero-uncertainty = true,
		mode                    = text,
	}\caption{Accuracy results for all real-world datasets when varying the strategy of the LEAP PE with fixed and
		learnable directions for different models.
The backbone architectures have around 4K parameters and we show the additional
		parameters each positional encoding introduces.
	}\label{tab:LEAP All Projections}
	\resizebox{\textwidth}{!}{
		\begin{tabular}{lll SS SS SS SS SS SS SS}
			\toprule
			                  &                          &                        & \multicolumn{2}{c}{\small\sc COX2} & \multicolumn{2}{c}{\small\sc BZR} & \multicolumn{2}{c}{\small\sc DHFR} & \multicolumn{2}{c}{\small\sc Letter-H} & \multicolumn{2}{c}{\small\sc Letter-M} & \multicolumn{2}{c}{\small\sc Letter-L} & \multicolumn{2}{c}{\small\sc Fingerprint}                                                                                                                                                                  \\
			\midrule
			{\small\sc Model} & {\small\sc Proj. Method} & {\small\sc Parameters} & {\small\sc \leaprd }               & {\small\sc \leapld }              & {\small\sc \leaprd }               & {\small\sc \leapld }                   & {\small\sc \leaprd }                   & {\small\sc \leapld }                   & {\small\sc \leaprd }                      & {\small\sc \leapld } & {\small\sc \leaprd } & {\small\sc \leapld } & {\small\sc \leaprd } & {\small\sc \leapld } & {\small\sc \leaprd } & {\small\sc \leapld } \\
			\midrule
			\multirow{5}{*}{\small\sc GCN}
			                  & Linear                   & {4K+2.5K}              & 79.2\pm0.6                         & 79.4\pm1.0                        & 78.8\pm0.6                         & 82.5\pm2.0                             & 70.9\pm3.2                             & 74.9\pm4.0                             & 72.2\pm3.3                                & 74.2\pm1.5           & 82.8\pm1.4           & 83.6\pm1.3           & 95.2\pm0.9           & 96.0\pm0.9           & 55.6\pm1.1           & 54.7\pm1.5           \\
			                  & Attn                     & {4K+5K  }              & 78.4\pm1.3                         & 79.0\pm0.9                        & 81.7\pm2.8                         & 82.5\pm1.6                             & 74.1\pm5.2                             & 77.3\pm4.1                             & 59.8\pm2.8                                & 62.7\pm2.6           & 74.4\pm1.5           & 73.9\pm4.6           & 92.3\pm1.3           & 94.5\pm1.4           & 53.0\pm1.9           & 54.4\pm1.2           \\
			                  & Attn PE                  & {4K+5K  }              & 78.2\pm1.2                         & 78.8\pm1.3                        & 82.5\pm2.4                         & 82.5\pm3.1                             & 73.2\pm3.7                             & 77.6\pm2.8                             & 67.2\pm1.5                                & 68.6\pm1.8           & 82.0\pm0.8           & 82.9\pm2.2           & 95.8\pm1.1           & 94.7\pm1.6           & 54.1\pm1.3           & 55.1\pm1.2           \\
			                  & DeepSets                 & {4K+.5K }              & 78.2\pm0.6                         & 79.0\pm1.2                        & 79.0\pm1.2                         & 81.7\pm3.5                             & 71.2\pm2.6                             & 73.3\pm3.6                             & 59.2\pm1.9                                & 63.4\pm2.1           & 72.4\pm0.6           & 76.0\pm1.1           & 91.4\pm1.8           & 92.0\pm0.6           & 52.3\pm1.3           & 54.1\pm1.3           \\
			                  & 1D Conv                  & {4K+.9K }              & 78.0\pm1.2                         & 79.2\pm2.0                        & 79.8\pm1.9                         & 82.5\pm2.7                             & 71.7\pm1.2                             & 76.7\pm3.4                             & 66.4\pm2.8                                & 63.1\pm3.2           & 81.6\pm1.5           & 81.7\pm3.5           & 94.2\pm1.5           & 93.4\pm1.4           & 55.6\pm1.1           & 54.0\pm2.2           \\
			\midrule
			\multirow{5}{*}{\small\sc GAT}
			                  & Linear                   & {4K+2.5K}              & 78.4\pm1.2                         & 79.7\pm2.0                        & 79.3\pm1.8                         & 81.7\pm2.4                             & 75.7\pm3.0                             & 76.3\pm2.2                             & 70.2\pm2.2                                & 73.5\pm2.1           & 82.4\pm2.3           & 82.4\pm1.6           & 95.8\pm0.8           & 95.1\pm0.9           & 55.1\pm0.6           & 54.8\pm2.1           \\
			                  & Attn                     & {4K+5K  }              & 78.4\pm0.5                         & 78.8\pm0.8                        & 82.0\pm3.4                         & 82.2\pm2.1                             & 74.9\pm3.1                             & 73.3\pm2.8                             & 62.0\pm1.2                                & 62.9\pm3.1           & 75.0\pm2.1           & 79.7\pm1.4           & 94.4\pm0.1           & 93.4\pm1.0           & 52.0\pm1.3           & 53.5\pm1.5           \\
			                  & Attn PE                  & {4K+5K  }              & 78.8\pm0.8                         & 80.1\pm2.2                        & 82.0\pm3.2                         & 79.5\pm0.7                             & 73.2\pm3.1                             & 75.9\pm5.8                             & 66.7\pm2.0                                & 65.0\pm3.5           & 83.2\pm1.1           & 79.3\pm2.4           & 94.0\pm1.3           & 95.1\pm1.1           & 54.5\pm1.8           & 54.5\pm1.8           \\
			                  & DeepSets                 & {4K+.5K }              & 79.2\pm1.6                         & 79.7\pm1.8                        & 81.2\pm2.0                         & 82.7\pm4.7                             & 71.2\pm4.1                             & 76.5\pm3.8                             & 58.8\pm2.6                                & 56.4\pm4.3           & 75.4\pm3.1           & 76.6\pm2.3           & 94.6\pm1.4           & 93.1\pm1.3           & 52.9\pm2.2           & 51.0\pm1.1           \\
			                  & 1D Conv                  & {4K+.9K }              & 78.4\pm0.9                         & 78.6\pm1.7                        & 81.7\pm4.0                         & 83.7\pm2.9                             & 70.6\pm2.3                             & 75.7\pm1.5                             & 67.3\pm2.0                                & 68.1\pm1.3           & 80.9\pm1.3           & 80.5\pm2.8           & 93.6\pm2.0           & 95.2\pm0.9           & 54.8\pm1.5           & 54.9\pm0.7           \\
			\midrule
			\multirow{5}{*}{\small\sc GIN}
			                  & Linear                   & {4K+2.5K}              & 78.4 \pm 0.8                       & 77.9 \pm 2.2                      & 79.3 \pm 1.4                       & 78.8 \pm 0.6                           & 76.2 \pm 3.2                           & 70.6 \pm 3.2                           & 62.7 \pm 3.4                              & 60.2 \pm 4.8         & 77.6 \pm 1.7         & 74.7 \pm 3.6         & 94.0 \pm 1.9         & 93.3 \pm 1.1         & 55.0 \pm 1.2         & 54.4 \pm 1.1         \\
			                  & Attn                     & {4K+5K  }              & 78.4 \pm 0.8                       & 78.2 \pm 0.5                      & 78.8 \pm 0.6                       & 78.5 \pm 1.9                           & 61.0 \pm 0.1                           & 73.9 \pm 4.1                           & 53.6 \pm 4.0                              & 46.3 \pm 8.1         & 73.2 \pm 5.2         & 65.2 \pm 9.4         & 92.8 \pm 1.3         & 89.6 \pm 0.8         & 55.3 \pm 1.4         & 53.9 \pm 0.9         \\
			                  & Attn PE                  & {4K+5K  }              & 78.2 \pm 0.5                       & 78.2 \pm 0.5                      & 78.8 \pm 0.6                       & 78.8 \pm 0.6                           & 61.0 \pm 0.1                           & 61.0 \pm 0.1                           & 58.4 \pm 4.8                              & 52.8 \pm 6.9         & 75.1 \pm 3.5         & 76.3 \pm 2.0         & 92.8 \pm 1.0         & 90.6 \pm 6.0         & 54.0 \pm 1.4         & 53.7 \pm 0.5         \\
			                  & DeepSets                 & {4K+.5K }              & 79.6 \pm 1.6                       & 79.0 \pm 0.9                      & 81.0 \pm 1.4                       & 80.2 \pm 2.0                           & 74.7 \pm 3.4                           & 69.2 \pm 2.2                           & 49.5 \pm 3.5                              & 53.6 \pm 6.9         & 71.2 \pm 2.0         & 67.4 \pm 4.6         & 89.5 \pm 1.9         & 87.2 \pm 3.8         & 54.5 \pm 0.7         & 53.3 \pm 0.5         \\
			                  & 1D Conv                  & {4K+.9K }              & 78.2 \pm 0.5                       & 78.4 \pm 1.4                      & 79.0 \pm 1.2                       & 81.2 \pm 1.4                           & 72.1 \pm 5.6                           & 70.1 \pm 6.8                           & 57.9 \pm 4.7                              & 52.7 \pm 2.7         & 76.5 \pm 3.7         & 72.7 \pm 6.3         & 91.8 \pm 1.7         & 90.6 \pm 0.8         & 53.8 \pm 0.4         & 54.2 \pm 1.1         \\
			\midrule
			\multirow{5}{*}{\small\sc NoMP}
			                  & Linear                   & {4K+2.5K}              & 79.0\pm0.6                         & 78.6\pm0.8                        & 83.2\pm1.7                         & 84.7\pm2.7                             & 74.3\pm6.1                             & 74.9\pm3.3                             & 79.5\pm1.2                                & 79.4\pm1.1           & 86.0\pm2.2           & 85.4\pm1.5           & 96.7\pm0.8           & 96.4\pm0.7           & 55.7\pm1.1           & 56.3\pm1.4           \\
			                  & Attn                     & {4K+5K  }              & 78.2\pm0.5                         & 78.2\pm0.5                        & 81.7\pm2.8                         & 79.0\pm0.9                             & 68.3\pm5.7                             & 71.7\pm4.1                             & 79.0\pm1.8                                & 81.3\pm1.9           & 84.8\pm0.9           & 86.5\pm2.6           & 96.1\pm0.6           & 97.2\pm0.8           & 53.8\pm0.7           & 54.8\pm1.2           \\
			                  & Attn PE                  & {4K+5K  }              & 78.4\pm0.4                         & 77.7\pm1.5                        & 78.8\pm0.6                         & 78.8\pm0.6                             & 64.2\pm4.8                             & 72.1\pm3.7                             & 81.3\pm1.9                                & 80.6\pm3.6           & 88.0\pm1.9           & 86.5\pm2.2           & 96.3\pm0.8           & 96.2\pm1.2           & 54.8\pm1.4           & 55.3\pm1.1           \\
			                  & DeepSets                 & {4K+.5K }              & 78.4\pm0.8                         & 78.0\pm0.5                        & 83.2\pm2.1                         & 82.0\pm2.8                             & 69.5\pm3.0                             & 72.6\pm3.6                             & 78.0\pm2.0                                & 79.2\pm1.9           & 86.0\pm2.2           & 87.5\pm2.0           & 96.3\pm0.6           & 96.6\pm0.8           & 54.0\pm0.4           & 54.3\pm1.0           \\
			                  & 1D Conv                  & {4K+.9K }              & 78.0\pm1.9                         & 78.1\pm1.1                        & 81.0\pm1.7                         & 79.3\pm1.8                             & 71.6\pm3.2                             & 75.7\pm2.7                             & 81.1\pm0.9                                & 81.6\pm1.9           & 87.0\pm2.0           & 88.5\pm2.5           & 97.2\pm0.3           & 98.0\pm0.4           & 54.5\pm1.0           & 54.1\pm0.4           \\
			\bottomrule
		\end{tabular}
	}
\end{table}

 \begin{figure}[h]
	\begin{subfigure}{\textwidth}
		\begin{subfigure}{0.32\textwidth}
			\includegraphics[width=\linewidth]{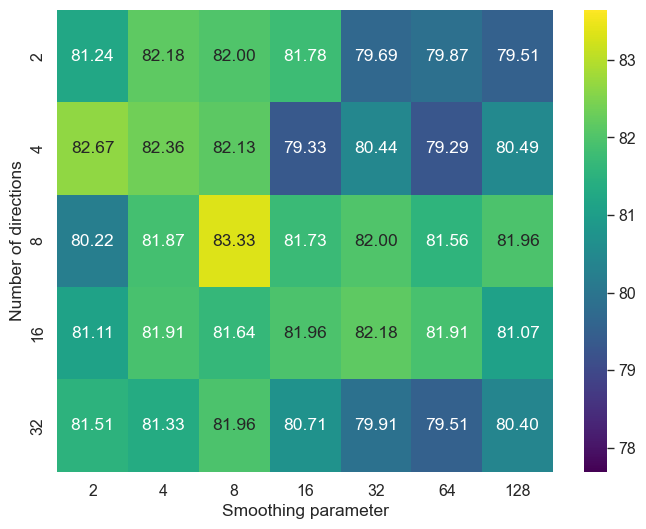}
			\caption*{Letter High}
		\end{subfigure}
		\begin{subfigure}{0.32\textwidth}
			\includegraphics[width=\linewidth]{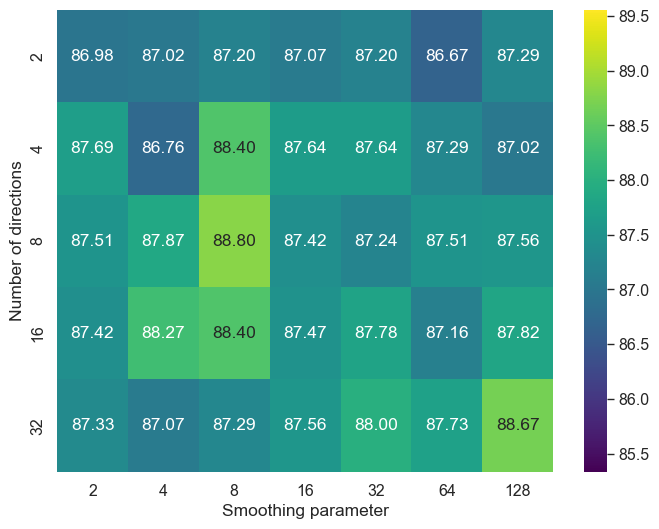}
			\caption*{Letter Medium}
		\end{subfigure}
		\begin{subfigure}{0.32\textwidth}
			\includegraphics[width=\linewidth]{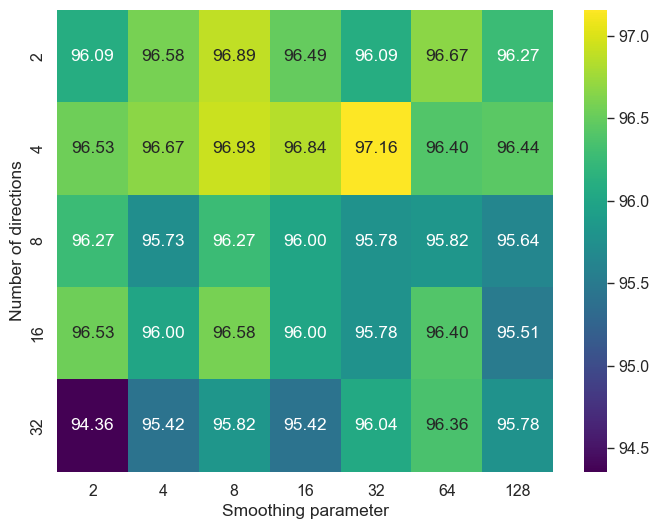}
			\caption*{Letter Low}
		\end{subfigure}
	\end{subfigure}
	\begin{subfigure}{\textwidth}
		\begin{subfigure}{0.32\textwidth}
			\includegraphics[width=\linewidth]{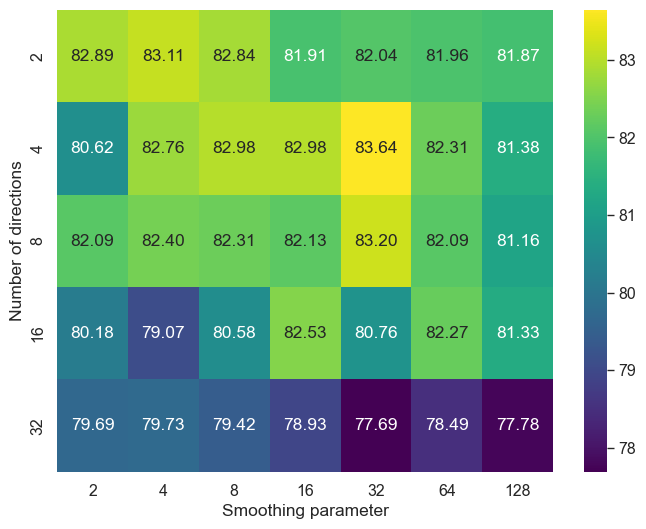}
			\caption*{Letter High}
		\end{subfigure}
		\begin{subfigure}{0.32\textwidth}
			\includegraphics[width=\linewidth]{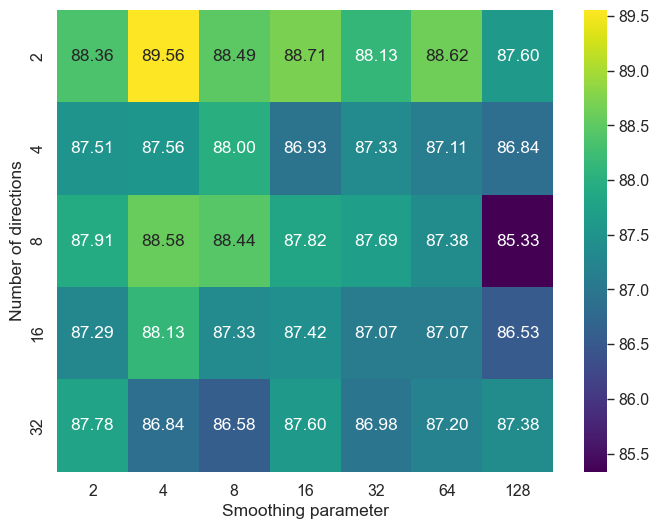}
			\caption*{Letter Medium}
		\end{subfigure}
		\begin{subfigure}{0.32\textwidth}
			\includegraphics[width=\linewidth]{images/ablations_fixed_low.png}
			\caption*{Letter Low}
		\end{subfigure}
\end{subfigure}
	\caption{We assess the sensitivity of LEAP with respect to the hyperparameters used in the ECT.
Top row shows the effect of changing the hyperparameters for LEAP-F (fixed directions)
		and the bottom row for LEAP-L (learnable directions).
LEAP consistently outperforms baselines across all settings and is thus robust with respect to the hyperparameters.
	}\label{fig:DECT Hyperparameter ablation}
\end{figure}
\begin{table}
	\centering
	\sisetup{
		detect-all              = true,
		table-format            = 2.1(1.1),
		detect-mode             = true,
		separate-uncertainty    = true,
		retain-zero-uncertainty = true,
		mode                    = text,
	}\caption{We provide a comparison with DECT~\citep{dect}. 
DECT summarizes the graph with a single global ECT and subsequently applies a convolutional 
        neural network for the classification. 
We compare our method with two variants of DECT, one with 4K parameters and one with 65K parameters.
The parameter count in LEAP ranges from 1K to 5K and therefore the comparison with DECT~(4K) 
        would be the most appropriate, although we outperform both variants on most datasets.
	}\begin{tabular}{l SSS SSS}
			\toprule
			{\small\sc Model}            & {\small\sc COX2} & {\small\sc BZR} & {\small\sc DHFR} & {\small\sc Letter-H} & {\small\sc Letter-M} & {\small\sc Letter-L} \\
			\midrule
			{\small\sc DECT (4K) }       & 70.4\pm 0.9      & 81.8\pm 3.2     & 67.9\pm 5.0      & 63.8\pm 6.0          & 76.2\pm 4.8          & 91.5\pm 2.1          \\
			{\small\sc DECT (65K)}       & 74.6\pm 4.5      & 84.3\pm 6.1     & 72.9\pm 1.6      & 85.4\pm 1.3          & 86.3\pm 2.0          & 96.8\pm 1.2          \\
			\midrule
			{\small\sc \leapld\,(GCN)}   & 79.4 \pm 1.0     & 82.5 \pm 1.6    & 77.6 \pm 2.8     & 74.2 \pm  1.5        & 83.6 \pm 1.3         & 96.0 \pm 0.9         \\
			{\small\sc \leapld\,(GAT)}   & 80.1 \pm 2.2     & 83.7 \pm 2.9    & 76.5 \pm 3.8     & 73.5 \pm 2.1         & 82.4 \pm 1.6         & 95.2 \pm 0.9         \\
			{\small\sc \leapld\,(NoMP) } & 78.6 \pm 0.8     & 84.7 \pm 2.7    & 75.7 \pm 2.7     & 81.6 \pm 1.9         & 88.5 \pm 2.5         & 98.0 \pm 0.4         \\
			\bottomrule
		\end{tabular}
\label{tab:LEAP DECT Comparison}
\end{table}
 
\clearpage
\section{Computational Performance}
\begin{table}[h]
	\centering
	\sisetup{
		detect-all              = true,
		table-format            = 2.2,
		detect-mode             = true,
		separate-uncertainty    = true,
		retain-zero-uncertainty = true,
		round-precision				  = 2,
		mode                    = text,
	}\caption{We report the average training time for the Letter High dataset.
The average training time measures the total training time, including potential early stopping.
For fair comparison between the various methods we also report the average training time per epoch.
LEAP yields both high performance while remaining fast to train.
	}\begin{tabular}{l SS }
			\toprule
			{\small\sc Method} & {\small\sc Training Time} & {\small\sc Average per Epoch}  \\
			\midrule
			{LaPE  }           & 60.35                         & 0.64                             \\
			{NoPE  }           & 44.60                         & 0.51                                                    \\
			{RWPE  }           & 35.77                         & 0.50                                                    \\
			{\leapld}          & 32.51                         & 0.93                                                    \\
			{\leaprd}          & 30.08                         & 0.84                                                   \\
			\bottomrule
		\end{tabular}
\end{table}

\begin{table}[h]
	\centering
	\sisetup{
		detect-all              = true,
		table-format            = 2.2,
		detect-mode             = true,
		separate-uncertainty    = true,
		retain-zero-uncertainty = true,
		round-precision				  = 2,
		mode                    = text,
	}\caption{For the Roman Empire dataset \citep{RomanEmpire} the average training time per epoch is reported as well as the initial preprocessing time. Since  this dataset consists in a single graph with 22662 nodes, 32927 edges, and 300-dimensional features, we choose it to evaluate how the runtimes of the PEs scale for larger graphs. We report preprocessing time as for this graph size it is \emph{not} negligible. Moreover, in this case, for the LEAP variants, the preprocessing consists in precomputing the node neighborhoods.  
LaPE requires the eigen decomposition of the full graph, leading to large initial preprocessing times.
LEAP remains computationally efficient.}\begin{tabular}{l SS }
			\toprule
			{\small\sc Method} & {\small\sc Training Time Per Epoch} & {\small\sc Preprocessing time}          \\
			\midrule
			{NoPE  }           & 0.09                             & 0.01   \\
			{LaPE  }           & 0.10                             & 164.83 \\
			{RWPE  }           & 0.10                             & 0.18   \\
			{\leaprd}          & 0.58                             & 11.57  \\
			{\leapld}          & 1.13                             & 11.57  \\
			\bottomrule
		\end{tabular}
\end{table}

\end{document}